# *Products of Weighted Logic Programs*∗

SHAY B. COHEN, ROBERT J. SIMMONS, and NOAH A. SMITH
*School of Computer Science*
*Carnegie Mellon University*
*E-mail:* {*scohen,rjsimmon,nasmith*}*@cs.cmu.edu*

**Abstract**

Weighted logic programming, a generalization of bottom-up logic programming, is a well-suited framework for specifying dynamic programming algorithms. In this setting, proofs correspond to the algorithm's output space, such as a path through a graph or a grammatical derivation, and are given a real-valued score (often interpreted as a probability) that depends on the real weights of the base axioms used in the proof. The desired output is a function over all possible proofs, such as a sum of scores or an optimal score. We describe the PRODUCT transformation, which can merge two weighted logic programs into a new one. The resulting program optimizes a product of proof scores from the original programs, constituting a scoring function known in machine learning as a "product of experts." Through the addition of intuitive constraining side conditions, we show that several important dynamic programming algorithms can be derived by applying PRODUCT to weighted logic programs corresponding to *simpler* weighted logic programs. In addition, we show how the computation of Kullback-Leibler divergence, an information-theoretic measure, can be interpreted using PRODUCT.

## 1 Introduction

Weighted logic programming is a technique that can be used to declaratively specify dynamic programming algorithms in a number of fields such as natural language processing (Manning and Schütze 1999) and computational biology (Durbin et al. 1998). Weighted logic programming is a generalization of bottom-up logic programming where each proof is assigned a score (or weight) that is a function of the scores of the axioms used in the proof. When these scores are interpreted as probabilities, then the solution to a whole weighted logic program can be interpreted in terms of probabilistic reasoning about unknowns, implying that the weighted logic program implements *probabilistic inference*.[1]

∗ To appear in Theory and Practice of Logic Programming (TPLP).

[1] The word *inference* has a distinct meaning in logic programming (e.g. "inference rule," "valid inference"), and so we will attempt to avoid confusion by using the *probabilistic* modifier whenever we are talking about probabilistic reasoning about unknowns.

Even though weighted logic programming is not limited to probabilistic inference, it is worth detailing their relationship. Let $I$, $A$, and $P$ be random variables, where the values of $I$ and $A$ are known and the value of $P$ is not known. Often there is a correspondence where

- $I$ corresponds to a conditional "input," encoded as axioms, known to be true;
- $A$ corresponds to a set of axioms known to be true; and
- $P$ corresponds to a deductive proof of the goal theorem using the axioms.

In the setting of weighted logic programming, there may be many different proofs of the goal given the set of axioms. We must therefore distinguish the weighted logic program from the "world" we are reasoning about in which these many different proofs of the goal correspond to different, mutually exclusive events, each of which has some probability of occurring. Weighted logic programming implements probabilistic inference over the value of the proof random variable $P$ given the values of $A$ and $I$: the weighted logic program implies a probability distribution $p(P \mid A, I)$, and it can be used to compute different useful quantities related to the distribution.

Previous work on weighted logic programming has shown that certain families of probabilistic models lend themselves extremely well to weighted logic programming as an inference mechanism. In general, weighted logic programming deals with probability distributions over objects with combinatorial structure—paths through graphs, grammatical derivations, and sequence alignments—that are quite useful in computer science applications.

In principle, one can think about combining such distributions with each other, creating distributions over even more complex structures that are related. This paper is about a natural extension to weighted logic programming as probabilistic inference over structures: combining weighted logic programs to perform inference over two or more structures. We describe a program transformation, PRODUCT, that implements joint probabilistic inference via weighted logic programming over two structured variables $P_1$ and $P_2$, when (a) each of the two separate structures can be independently reasoned about using weighted logic programming, and (b) the joint model factors into a product of two distributions $p(P_1 \mid A_1, I_1)$ and $p(P_2 \mid A_2, I_2)$.[2]

As a program transformation on traditional logic programs, PRODUCT is not novel; it has existed as a compiler transformation for over a decade (Pettorossi and Proietti 1994; Pettorossi 1999). As a way of describing joint probabilistic inference in weighted logic programming, the transformation has been intuitively exploited in designing algorithms for specific applications, but has not, to our knowledge, been generalized. The contribution of this paper is a general, intuitive, formal setting for dynamic programming algorithms that process two or more conceptually distinct objects. Indeed, we show that many important dynamic programming algorithms can be derived using simpler "factor" programs and the PRODUCT transformation together with side conditions that capture the relationship between the structures.

The paper is organized as follows. In §2 we give an overview of weighted logic

[2] In the language of probability, this means that $P_1$ and $P_2$ are conditionally independent given $A_1$, $A_2$, $I_1$, and $I_2$.

$$\texttt{reachable(Q)} \;\; \text{:-} \;\; \texttt{initial(Q)}. \tag{1}$$

$$\texttt{reachable(Q)} \;\; \text{:-} \;\; \texttt{reachable(P)}, \texttt{edge(P, Q)}. \tag{2}$$

Fig. 1. A simple bottom-up logic program for graph reachability.

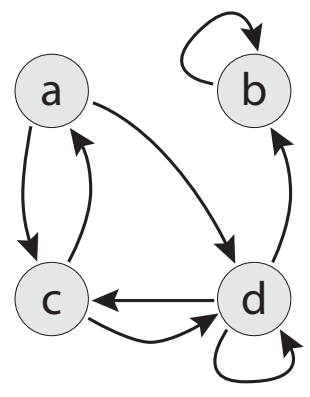

| | | | | | |
|---|---|---|---|---|---|
| initial(a) | = | T | edge(c, d) | = | T |
| edge(a, c) | = | T | edge(d, b) | = | T |
| edge(a, d) | = | T | edge(d, c) | = | T |
| edge(b, b) | = | T | edge(d, d) | = | T |
| edge(c, a) | = | T | | | |

Fig. 2. A directed graph and the corresponding initial database.

programming. In §3 we describe products of experts, a concept from machine learning that elucidates the kinds of probabilistic models amenable to our framework. In §4 we describe the PRODUCT transformation. In §5 we give show how several well-known algorithms can be derived using the PRODUCT transformation applied to simpler algorithms. §6 presents some variations on the PRODUCT transformation. In §7 we show how to use the PRODUCT transformation and a specially designed semiring to calculate important information theoretic quantities related to probability distributions over proofs.

## 2 Weighted Logic Programming

To motivate weighted logic programming, we begin with a logic program for single-source connectivity on a directed graph, shown in Figure 1. In the usual bottom-up interpretation of this program, an initial database (i.e., set of axioms) would describe the edge relation and one (or more) starting vertices as axioms of the form initial(a) for some a. Repeated forward inference can then be applied on the rules in Figure 1 to find the least database closed under those rules. However, in traditional logic programming this program can *only* be understood as a program calculating connectivity over a graph.

Weighted logic programming generalizes traditional logic programming. In traditional logic programming, a proof is a tree of valid (deductive) inferences from axioms, and a valid atomic proposition is one that has at least one proof. In weighted logic programming we generalize this notion: axioms, proofs, and atomic propositions are said to “have values” rather than just “be valid.” Traditional logic programs can be understood as weighted logic programs with Boolean values: axioms all have the value “true,” as do all valid propositions. The single-source connectivity program would describe the graph in Figure 2 by assigning T as the value of all the existing edges and the proposition initial(a).

### *2.1 Non-Boolean Programs*

With weighted logic programming, the axioms and propositions can be understood as having non-Boolean values. In Figure 3, each axiom of the form $\mathtt{edge}(\mathtt{X},\mathtt{Y})$ is given a value corresponding to the cost associated with that edge in the graph, and the axiom $\mathtt{initial}(\mathtt{a})$ is given the value 0. If we take the value or "score" of a proof to be the the *sum* of the values the axioms at its leaves and take the value of a proposition to be the *minimum* score over all possible proofs, then the program from Figure 1 gives a declarative specification of the *single-source shortest path* problem. Multiple uses of an axiom in a proof are meaningful: if a proof includes the $\mathtt{edge}(\mathtt{d},\mathtt{d})$ axiom once, it corresponds to a single traversal of the loop from d to d and adds a cost of 2, and if a proof includes the axiom twice, it corresponds to two distinct traversals and adds a cost of 4.

We replace the connectives :- (disjunction) and , (conjunction) with $\min =$ and $+$, respectively, and interpret the WLP over the non-negative numbers. With a specific execution model, the result is Dijkstra's single-source shortest-path algorithm.

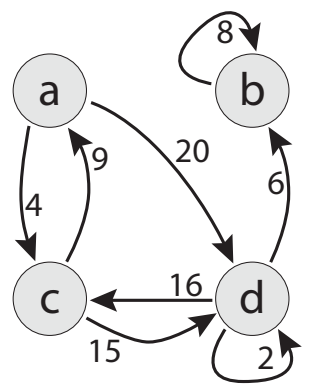

| | | | | | |
|---|---|---|---|---|---|
| $\mathtt{initial}(\mathtt{a})$ | = | 0 | $\mathtt{edge}(\mathtt{c},\mathtt{d})$ | = | 15 |
| $\mathtt{edge}(\mathtt{a},\mathtt{c})$ | = | 4 | $\mathtt{edge}(\mathtt{d},\mathtt{b})$ | = | 6 |
| $\mathtt{edge}(\mathtt{a},\mathtt{d})$ | = | 20 | $\mathtt{edge}(\mathtt{d},\mathtt{c})$ | = | 16 |
| $\mathtt{edge}(\mathtt{b},\mathtt{b})$ | = | 8 | $\mathtt{edge}(\mathtt{d},\mathtt{d})$ | = | 2 |
| $\mathtt{edge}(\mathtt{c},\mathtt{a})$ | = | 9 | | | |

Fig. 3. A cost graph and the corresponding initial database.

In addition to the cost-minimization interpretation in Figure 3, we can interpret weights on edges as *probabilities* and restate the problem in terms of probability *maximization.* In Figure 4, the outgoing edges from each vertex sum to at most 1. If we assign the missing 0.1 probability from vertex b to a "stopping" event—either implicitly or explicitly by modifying the axioms—then each vertex's outgoing edges sum to exactly one and the graph can be seen as a Markov model or probabilistic finite-state network over which random walks are well-defined. If we replace the connectives :- (disjunction) and , (conjunction) with $\max =$ and $\times$, then the value of $\mathtt{reachable}(\mathtt{X})$ for any X is the probability of the most likely path from a to X. For instance, $\mathtt{reachable}(\mathtt{a})$ ends up with the value 1, and $\mathtt{reachable}(\mathtt{b})$ ends up with value 0.16, corresponding to the path $\mathtt{a} \to \mathtt{d} \to \mathtt{b}$, whose weight is (value of $\mathtt{initial}(\mathtt{a})$ $\times$ value of $\mathtt{edge}(\mathtt{a},\mathtt{d})$ $\times$ value of $\mathtt{edge}(\mathtt{d},\mathtt{b})$).

If we keep the initial database from Figure 4 but change our operators from $\max =$ and $\times$ to $+=$ and $\times$, the result is a program for *summing* over the probabilities of all distinct paths that start in a and lead to X, for each vertex X. This quantity is known as the "path sum" (Tarjan 1981). The path sum for $\mathtt{reachable}(\mathtt{b})$, for

$$\texttt{reachable(Q)} \ \oplus\!\!= \ \texttt{initial(Q)}. \tag{3}$$

$$\texttt{reachable(Q)} \ \oplus\!\!= \ \texttt{reachable(P)} \otimes \texttt{edge(P,Q)}. \tag{4}$$

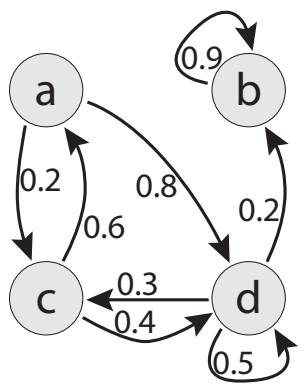

| | | | | | |
|---|---|---|---|---|---|
| `initial(a)` | = | 1 | `edge(c,d)` | = | 0.4 |
| `edge(a,c)` | = | 0.2 | `edge(d,b)` | = | 0.2 |
| `edge(a,d)` | = | 0.8 | `edge(d,c)` | = | 0.3 |
| `edge(b,b)` | = | 0.9 | `edge(d,d)` | = | 0.5 |
| `edge(c,a)` | = | 0.6 | | | |

Fig. 4. A probabilistic graph and the corresponding initial database. With stopping probabilities made explicit, this would encode a Markov model.

Fig. 5. The logic program from Figure 1, rewritten to emphasize that it is generalized to an arbitrary semiring.

instance, is 10—this is not a probability, but rather an infinite sum of probabilities of many paths, some of which are prefixes of each other.[3]

These three related weighted logic programs are useful generalizations of the reachability logic program in Figure 1. Figure 5 gives a generic representation of all four algorithms in the Dyna language (Eisner et al. 2005). The key difference among them is the *semiring* in which we interpret the weights. An algebraic semiring consists of five elements $\langle \mathbb{K}, \oplus, \otimes, \mathbf{0}, \mathbf{1} \rangle$, where $\mathbb{K}$ is a domain closed under $\oplus$ and $\otimes$, $\oplus$ is a binary, associative, commutative operator, $\otimes$ is a binary, associative operator that distributes over $\oplus$, $\mathbf{0} \in \mathbb{K}$ is the $\oplus$-identity, and $\mathbf{1} \in \mathbb{K}$ is the $\otimes$-identity. We require, following Goodman (1999), that the semirings we use be **complete**. Complete semirings are semirings with the additional property that they are closed under finite products and infinite sums—in our running example, this corresponds to the idea that there may be infinitely many paths through a graph, all with finite length. Complete semirings also have the property that infinite sums behave like finite ones—they are associative and commutative, and the multiplicative operator distributes over them.

In our running example, reachability uses the Boolean semiring $\langle \{\text{T}, \text{F}\}, \vee, \wedge, \text{F}, \text{T} \rangle$, single-source shortest path uses $\langle \mathbb{R}_{\geq 0} \cup \{\infty\}, \min, +, \infty, 0 \rangle$, the most-probable-path variant uses $\langle [0,1], \max, \times, 0, 1\rangle$, and the probabilistic path-sum variant uses the so-called "real" semiring $\langle \mathbb{R}_{\geq 0} \cup \{\infty\}, +, \times, 0, 1\rangle$.

[3] Clearly "10" is not a meaningful probability, but that is a result of the loop from `b` to `b` with probability 0.9—in fact, one informal way of looking at the result is simply to observe that $10 = 1 + 0.9 + (0.9)^2 + (0.9)^3 + \ldots$, corresponding to proofs of `reachable(b)` that include `edge(b,b)` zero, one, two, three, ... times. If we added an axiom `edge(b,final)` with weight 0.1 representing the 10% probability of stopping at any step in state `b`, then the path sum for `reachable(final)` would be $10 \times 0.1 = 1$, which is a reasonable probability that corresponds to the fact that a graph traversal can be arbitrarily long but has a 100% chance of eventually reaching `b` and then stopping.

Weighted logic programming developed primarily within the computational linguistics community. Building upon the observations of Shieber, Schabes, and Pereira (1995) and Sikkel (1997) that many parsing algorithms for nondeterministic grammars could be represented as deductive logic programs, Goodman (1999) showed that the structure of the parsing algorithms was amenable to interpretation on a number of semirings. McAllester (1999) additionally showed that this representation facilitates reasoning about asymptotic complexity. Other developments include a connection between weighted logic programs and hypergraphs (Klein and Manning 2004), optimal A* search for maximizing programs (Felzenszwalb and McAllester 2007), semiring-general agenda-based implementations (Eisner et al. 2005), improved $k$-best algorithms (Huang and Chiang 2005), and program transformations to improve efficiency (Eisner and Blatz 2007).

## *2.2 Formal Definition*

Eisner and Blatz (2007) describe the semantics of weighted logic programs in detail; we summarize their discussion in this section and point the reader to that paper for further detail. A weighted logic program is a set of *Horn equations* describing a set of declarative, usually recursive equations over an abstract semiring. Horn equations, which we will refer to by the shorter and more traditional term *rules*, take the form

$$\texttt{consequent}(\mathbf{U}) \oplus\!= \texttt{antecedent}_1(\mathbf{W_1}) \otimes \cdots \otimes \texttt{antecedent}_n(\mathbf{W_n}).$$

Here $\mathbf{U}$ and the $\mathbf{W_i}$ are sequences of terms which include free variables. If the variables in $\mathbf{U}$ are a subset of the variables in $\mathbf{W_1} \ldots \mathbf{W_n}$ for every rule, then the program is *range restricted* or *fully grounded*.

We can also give rules *side conditions*. Side conditions are additional constraints that are added to a rule to remove certain proofs from consideration. For example, side conditions could allow us to modify rule 4 in Figure 5 to disallow self-loops and only allow traversal of an edge when there was another edge in the opposite direction:

$$\texttt{reachable}(\texttt{Q}) \quad \oplus\!= \quad \texttt{reachable}(\texttt{P}) \otimes \texttt{edge}(\texttt{P},\texttt{Q}) \texttt{ if } \texttt{edge}(\texttt{Q},\texttt{P}) \wedge \texttt{Q} \neq \texttt{P}. \qquad (5)$$

Side conditions do not change the value of any individual proof, they only filter out any proof that does not satisfy the side conditions. In this paper, we use mostly side conditions that enforce equality between variables. For a more thorough treatment of side conditions see Goodman (1999) or Eisner and Blatz (2007).

A weighted logic program is specified on an arbitrary semiring, and can be interpreted in any semiring $\langle \mathbb{K}, \oplus, \otimes, \mathbf{0}, \mathbf{1}\rangle$ as previously described. The meaning of a weighted logic program is determined by the rules together with a set of fully grounded axioms (or *facts* in the Prolog setting). Each axiom is assigned a value from the set $\mathbb{K}$ that is interpreted as a weight or score.

A common idiom in weighted logic programming is to specify the query as a distinguished predicate `goal` that takes no arguments. A computationally uninteresting (because are no intermediate computation steps) but otherwise legitimate way to present a weighted logic program is as a single rule of the form

$$\texttt{goal} \quad \oplus\!= \quad \texttt{axiom}_\texttt{1}(\mathbf{W_1}) \otimes \cdots \otimes \texttt{axiom}_\texttt{n}(\mathbf{W_n}).$$

In this degenerate case, each distinct way of satisfying the premises using axioms in the database would correspond to a distinct proof of `goal`. The score of each proof would be given by the semiring-product of the scores of the axioms, and the value of `goal` would be determined by the semiring-sum of the scores of all the proofs.

In the general case, the value of the proposition/theorem `goal` is a semiring-sum over all of its proofs, starting from the axioms, where the value of any single proof is the semiring-product of the axioms involved. This is effectively encoded using the inference rules as a sum of products of sums of products of ... sums of products, exploiting distributivity and shared substructure for efficiency. This inherent notion of shared substructure means that weighted logic programming can give straightforward declarative specifications for problems that are typically solved by dynamic programming. The Dyna programming language implements a particular dynamic programming strategy for implementing these declarative specifications (Eisner et al. 2005), though the agenda algorithm that it implements may potentially have significantly different behavior, in terms of time and space complexity, than other dynamic programming algorithms that meet the same specification.

In many practical applications, as in our reachability example in §2.1, values are interpreted as probabilities to be maximized or summed or costs to be minimized.

## 3 Weighted Logic Programs and Probabilistic Reasoning

In this section, we will return focus to the probabilistic interpretation of weighted logic programs that we first described in the introduction. In §3.1 we will describe in more detail how the results of weighted logic programs are interpreted as probabilities—readers with a background in statistics and machine learning can probably skip or skim this section. In §3.2, we will introduce the notion of a *product of experts* that motivates the PRODUCT transformation.

Our running example for this section is a probabilistic finite-state automaton

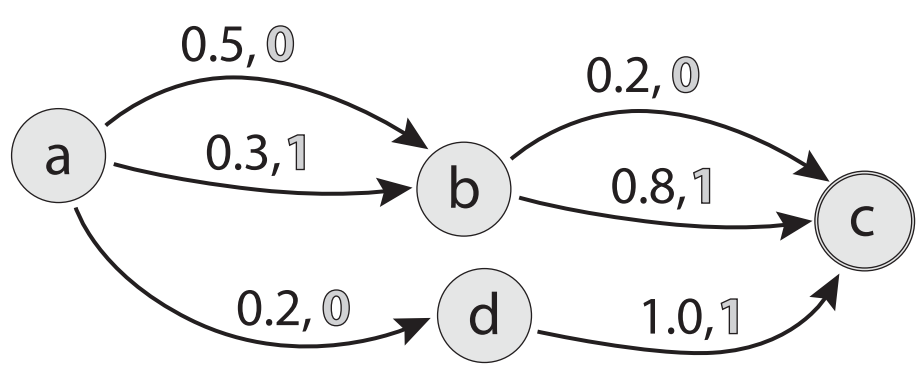

| | | |
|---|---|---|
| `initial(a)` | = | 1 |
| `final(c)` | = | 1 |
| `arc(a, b, 0)` | = | 0.5 |
| `arc(a, b, 1)` | = | 0.3 |
| `arc(a, d, 0)` | = | 0.2 |
| `arc(b, c, 0)` | = | 0.2 |
| `arc(b, c, 1)` | = | 0.8 |
| `arc(d, c, 1)` | = | 1.0 |

Fig. 6. A probabilistic FSA and the corresponding initial database.

$$\texttt{goal} \oplus\!= \texttt{path}(\texttt{Q}) \otimes \texttt{final}(\texttt{Q}). \quad (6)$$
$$\texttt{path}(\texttt{Q}) \oplus\!= \texttt{initial}(\texttt{Q}). \quad (7)$$
$$\texttt{path}(\texttt{Q}) \oplus\!= \texttt{path}(\texttt{P}) \otimes \texttt{arc}(\texttt{P}, \texttt{Q}, \texttt{A}). \quad (8)$$

Fig. 7. The weighted logic program for weighted FSA reachability.

$$\texttt{goal} \oplus\!= \texttt{path}(\texttt{Q}, \texttt{I}) \otimes \texttt{final}(\texttt{Q}) \otimes \texttt{length}(\texttt{I}). \quad (9)$$
$$\texttt{path}(\texttt{Q}, \texttt{0}) \oplus\!= \texttt{initial}(\texttt{Q}). \quad (10)$$
$$\texttt{path}(\texttt{Q}, \texttt{I}) \oplus\!= \texttt{path}(\texttt{P}, \texttt{I} - 1) \otimes \texttt{arc}(\texttt{P}, \texttt{Q}, \texttt{A}) \otimes \texttt{string}(\texttt{I}, \texttt{A}). \quad (11)$$

Fig. 8. The weighted logic program for weighted FSA recognition.

over the alphabet $\{\texttt{0}, \texttt{1}\}$, shown in Figure 6. The most probable path through the graph is the one that recognizes the string "`01`" by going through states `a`, `b`, and `c`, and that the probability of this path is 0.4. Other than the labels on the edges, this is the same setup used in the graph-reachability example from Figure 4. The `edge` predicate from the previous section is now called `arc` and has been augmented to carry a third argument representing an output character.

### 3.1 The Probabilistic Interpretation of Weighted Logic Programming

Recall from the introduction that, in the context of weighted logic programming, we have random variables $I$, $A$, and $P$, where

- $I$ corresponds to a set of conditional "input" axioms known to be true;
- $A$ corresponds to a set of axioms known to be true; and
- $P$ corresponds to a deductive proof of the goal theorem using the axioms.

In this case, $I$ corresponds to one of the various possible sentences recognized by the FSA (i.e., `00`, `01`, `10`, and `11`). $A$ corresponds to a particular directed graph with weighted edges, encoded by a set of axioms. $P$ corresponds to an individual proof/path through the graph. In Figure 7, which is the straightforward adaptation of the reachability program in Figure 5 to labeled edges, the value of `goal` in the most-probable-path semiring is $\max_{\mathit{proof}} p(P = \mathit{proof}, I = \mathit{sentence} \mid A = \mathit{graph})$—the value of the most probable path emitting *any* possible sentence $I$.

In order to talk about the input sentences $I$, we first add a set of axioms that describe $I$. If we are interested in the sentence "`01`" we would add axioms $\texttt{string}(1, \texttt{0})$, $\texttt{string}(2, \texttt{1})$, and $\texttt{length}(2)$, whereas if we were interested in the sentence "`hey`" we would add axioms $\texttt{string}(1, \texttt{h})$, $\texttt{string}(2, \texttt{e})$, $\texttt{string}(3, \texttt{y})$, and $\texttt{length}(3)$. These axioms are all given the value 1 (the multiplicative unit of the semiring), and so they could equivalently be treated as side conditions. With these new axioms, we modify

Figure 7 to obtain Figure 8, a weighted logic program that limits the proofs/paths to the ones which represent recognition of the input string $I$.[4]

Now, Figure 8 interpreted over the most-probable-path semiring *does* allow us to find the proof that, given the edge weights and a specific sentence, maximizes $p(P = \mathit{proof} \mid I = \mathit{sentence}, A = \mathit{graph})$. It does *not*, however, give us $p(P = \mathit{proof} \mid I = \mathit{sentence}, A = \mathit{graph})$, but rather $p(P = \mathit{proof}, I = \mathit{sentence} \mid A = \mathit{graph})$, the *joint* probability of a path and a sentence given the weights on the edges.

Concretely, in our running example there are five possible proofs of `goal` in Figure 7 whose probabilities sum to 1, but there are only two parses that also recognize the string "`01`," which are $\mathtt{a}_0\mathtt{b}_1\mathtt{c}$ with weight 0.4 and $\mathtt{a}_0\mathtt{d}_1\mathtt{c}$ with weight 0.2—the route through `b` is twice as likely. The value of `goal` in Figure 8 interpreted in the most-probable-path semiring would be 0.4 (the joint probability of obtaining the proof $\mathtt{a}_0\mathtt{b}_1\mathtt{c}$ and of recognizing the string "`01`") not $0.\overline{6}$ (the probability of the proof $\mathtt{a}_0\mathtt{b}_1\mathtt{c}$ given the sentence "`01`"). In other words, we have: $p(P = \mathtt{a}_0\mathtt{b}_1\mathtt{c}, I = \mathtt{01} \mid A = \text{Fig. 6}) = 0.4$, $p(P = \mathtt{a}_0\mathtt{d}_1\mathtt{c}, I = \mathtt{01} \mid A = \text{Fig. 6}) = 0.2$, $p(P = \mathtt{a}_0\mathtt{b}_1\mathtt{c} \mid I = \mathtt{01}, A = \text{Fig. 6}) = 0.\overline{6}$.

The solution for correctly discovering the conditional probability lies in the fact that the joint and conditional probabilities are related in the following way:

$$p(P \mid A, I) = \frac{p(P, I \mid A)}{p(I \mid A)}$$

This, combined with the knowledge that the *marginal* probability $p(I \mid A)$ is the result of evaluating Figure 8 over the path-sum semiring (i.e., $\langle \mathbb{R}_{\geq 0} \cup \{\infty\}, +, \times, 0, 1\rangle$), allows us to correctly calculate not only the most probable proof $P$ of a given sentence but also the probability of that proof *given* the sentence. The marginal probability in our running example is 0.6, and $0.4/0.6 = 0.\overline{6}$, which is the desired result.

To restate this in a way that is more notationally consistent with other work in machine learning, we first take the weighted axioms $A$ as implicit. Then, instead of proofs $P$ we talk about values $y$ for a random variable $Y$ drawn out of a domain $\mathcal{Y}$ (the space of possible structures, which in our setting corresponds to the space of possible proofs), and instead of inputs $I$ we talk about values $x$ for a random variable $X$ drawn out of a domain $\mathcal{X}$ (the space of all possible inputs).

Then, to predict the most likely observed value for $y$, denoted $\hat{y}$, we have the following formula:

$$\hat{y} = \operatorname*{argmax}_{y \in \mathcal{Y}} p(Y = y \mid X = x) = \operatorname*{argmax}_{y \in \mathcal{Y}} \frac{p(Y = y, X = x)}{p(X = x)} \tag{12}$$

Because $p(X = x)$ does not depend on $y$, if we only want to know $\hat{y}$ it suffices to find the $y$ that maximizes $p(Y = y, X = x)$ (which was written as $p(P = \mathit{proof}, I =$

[4] Rule 11 in this figure uses "$\mathtt{I} - \mathtt{1}$" in a premise: we assume that our formalism includes natural numbers that support increment/decrement operations, and our simple uses can be understood as syntactic shorthand either for structured terms (`z`, `s(z)`, etc.) or for the use of primitive side conditions such as $\mathtt{inc}(\mathtt{I}, \mathtt{I}')$.

*sentence* | $A =$ *axioms*) above). One way to do this is to execute a weighted logic program in the most-probable-path semiring.

### 3.2 Products of Experts

Of recent interest are probability models $p$ that take a *factored form*, for example:

$$p(Y = y \mid X = x) \; \propto \; p_1(Y = y \mid X = x) \times \cdots \times p_n(Y = y \mid X = x) \tag{13}$$

where $\propto$ signifies "proportional to" and suppresses the means by which the probability distribution is renormalized to sum to one. This kind of model is called a *product of experts* (Hinton 2002). Intuitively, the probability of an event under $p$ can only be relatively large if "all the experts concur," i.e., if the probability is large under each of the $p_i$. Any single expert can make an event arbitrarily unlikely (even impossible) by giving it very low probability, and the solution to Equation 12 for a product of experts model will be the $y \in \mathcal{Y}$ (here, a proof) least objectionable to all experts.

The attraction of such probability distributions is that they modularize complex systems (Klein and Manning 2003; Liang et al. 2008). They can also offer computational advantages when solving Equation 12 (Chiang 2007). Further, the expert factors can often be trained (i.e., estimated from data) separately, speeding up expensive but powerful machine learning methods (Smith and Smith 2004; Sutton and McCallum 2005; Smith et al. 2005; Cohen and Smith 2007).

To the best of our knowledge, there has been no attempt to formalize the following intuitive idea about products of experts: algorithms for reasoning about mutually constrained product proof values should resemble the individual algorithms for each of the two separate "factor" proofs' values. Our formalization is intended to aid in algorithm development as new kinds of complex random variables are coupled, with a key practical advantage: the expert factors are known because they fundamentally *underlie* the main algorithm. Indeed, we call our algorithms "products" because they are derived from "factors," analogous to the product of expert probability models that are derived from factor expert probability models.

To relate this observation to the running example from this section, imagine we created two copies of Figure 8 which operated over the same sentence (as described by `string` and `length` predicates) but which had different predicates and axioms $\texttt{goal}_1$, $\texttt{path}_1$, $\texttt{final}_1$, $\texttt{initial}_1$, and $\texttt{arc}_1$ (and likewise $\texttt{goal}_2$, $\texttt{path}_2$, etc.). Consider a combined goal predicate $\texttt{goal}_{1\bullet 2}$ defined by the rule

$$\texttt{goal}_{1\bullet 2} \quad \oplus\!= \quad \texttt{goal}_1 \otimes \texttt{goal}_2. \tag{14}$$

Now we have two experts ($\texttt{goal}_1$ and $\texttt{goal}_2$), and we literally take the (semiring) product of them, but this is still not quite the "product of experts," because the proofs of the goals are allowed to be independent. In other words, what we have is the following:

$$p(Y_1 = y_1, Y_2 = y_2 \mid X = x) \; \propto \; p_1(Y_1 = y_1 \mid X = x) \times p_2(Y_2 = y_2 \mid X = x).$$

The PRODUCT transformation is a meaning-preserving transformation on weighted logic programs that exposes the joint structure in such a way that—depending on our domain-specific understanding of what it means for the two proofs $y_1$ and $y_2$ to match—allows us to add constraints that result in a weighted logic program that forces the structures to match, as required by the specification in Equation 13.

## 4 Products of Weighted Logic Programs

In this section, we will motivate products of weighted logic programs in the context of the running example of generalized graph reachability. We will then define the PRODUCT transformation precisely and describe the process of specifying new algorithms as constrained versions of product programs.

The PRODUCT transformation can be seen as an instance of the *tupling* program transformation combined with an *unfold/fold* transformation (Pettorossi and Proietti 1994; Pettorossi 1999) that preserves the meaning of programs. However, we are interested in this transformation not for reasons of efficiency, but because it has the effect of exposing the shared structure of the two individual programs in such a way that, by the manual addition of constraints, we can force the two original programs to optimize over the *same* structures, thereby implementing optimization over the product of experts as described in the previous section. The addition of constraints requires an understanding of the problem at hand, as we show in §5 by presenting a number of examples.

### *4.1 The Product of Graph Reachability Experts*

Figure 9 defines two experts, copies of the graph-reachability program from Figure 5. We are interested in a new predicate $\texttt{reachable}_{1\bullet2}(\texttt{Q}_1,\texttt{Q}_2)$, which for any particular $\texttt{Q}_1$ and $\texttt{Q}_2$ should be equal to the product of $\texttt{reachable}_1(\texttt{Q}_1)$ and $\texttt{reachable}_2(\texttt{Q}_2)$. Just as we did in our thought experiment with $\texttt{goal}_{1\bullet2}$ in the previous section, we could define the predicate by adding the following rule to the program in Figure 9:

$$\texttt{reachable}_{1\bullet2}(\texttt{Q}_1,\texttt{Q}_2) \quad \oplus= \quad \texttt{reachable}_1(\texttt{Q}_1) \otimes \texttt{reachable}_2(\texttt{Q}_2).$$

This program is a bit simplistic, however; it merely describes calculating the experts independently and then combining them at the end.

The predicate $\texttt{reachable}_{1\bullet2}$ can alternatively be calculated by adding the following four rules to Figure 9:

$$\begin{array}{lcl}
\texttt{reachable}_{1\bullet2}(\texttt{Q}_1,\texttt{Q}_2) & \oplus= & \texttt{initial}_1(\texttt{Q}_1) \otimes \texttt{initial}_2(\texttt{Q}_2). \\
\texttt{reachable}_{1\bullet2}(\texttt{Q}_1,\texttt{Q}_2) & \oplus= & \texttt{initial}_1(\texttt{Q}_1) \otimes \texttt{reachable}_2(\texttt{P}_2) \otimes \texttt{edge}_2(\texttt{P}_2,\texttt{Q}_2). \\
\texttt{reachable}_{1\bullet2}(\texttt{Q}_1,\texttt{Q}_2) & \oplus= & \texttt{reachable}_1(\texttt{P}_1) \otimes \texttt{edge}_1(\texttt{P}_1,\texttt{Q}_1) \otimes \texttt{initial}_2(\texttt{Q}_2). \\
\texttt{reachable}_{1\bullet2}(\texttt{Q}_1,\texttt{Q}_2) & \oplus= & \texttt{reachable}_1(\texttt{P}_1) \otimes \texttt{edge}_1(\texttt{P}_1,\texttt{Q}_1) \otimes \\
 & & \texttt{reachable}_2(\texttt{P}_2) \otimes \texttt{edge}_2(\texttt{P}_2,\texttt{Q}_2).
\end{array}$$

This step is described as an *unfold* by Pettorossi (1999). This unfold can then be followed by a *fold*: because $\texttt{reachable}_{1\bullet2}(\texttt{Q}_1,\texttt{Q}_2)$ was defined above to be the product of $\texttt{reachable}_1(\texttt{Q}_1)$ and $\texttt{reachable}_2(\texttt{Q}_2)$, we can replace each instance of the two premises $\texttt{reachable}_1(\texttt{Q}_1)$ and $\texttt{reachable}_2(\texttt{Q}_2)$ with the single premise $\texttt{reachable}_{1\bullet2}(\texttt{Q}_1,\texttt{Q}_2)$.

$$\texttt{reachable}_1(\texttt{Q}_1) \oplus= \texttt{initial}_1(\texttt{Q}_1). \quad (15)$$
$$\texttt{reachable}_1(\texttt{Q}_1) \oplus= \texttt{reachable}_1(\texttt{P}_1) \otimes \texttt{edge}_1(\texttt{P}_1, \texttt{Q}_1). \quad (16)$$

$$\texttt{reachable}_2(\texttt{Q}_2) \oplus= \texttt{initial}_2(\texttt{Q}_2). \quad (17)$$
$$\texttt{reachable}_2(\texttt{Q}_2) \oplus= \texttt{reachable}_2(\texttt{P}_2) \otimes \texttt{edge}_2(\texttt{P}_2, \texttt{Q}_2). \quad (18)$$

Fig. 9. Two identical experts for generalized graph reachability, duplicates of the program in Figure 5.

$$\texttt{reachable}_{1\bullet 2}(\texttt{Q}_1, \texttt{Q}_2) \oplus= \texttt{initial}_1(\texttt{Q}_1) \otimes \texttt{initial}_2(\texttt{Q}_2). \quad (19)$$
$$\texttt{reachable}_{1\bullet 2}(\texttt{Q}_1, \texttt{Q}_2) \oplus= \texttt{reachable}_2(\texttt{P}_2) \otimes \texttt{edge}_2(\texttt{P}_2, \texttt{Q}_2) \otimes \texttt{initial}_1(\texttt{Q}_1). \quad (20)$$
$$\texttt{reachable}_{1\bullet 2}(\texttt{Q}_1, \texttt{Q}_2) \oplus= \texttt{reachable}_1(\texttt{P}_1) \otimes \texttt{edge}_1(\texttt{P}_1, \texttt{Q}_1) \otimes \texttt{initial}_2(\texttt{Q}_2). \quad (21)$$
$$\texttt{reachable}_{1\bullet 2}(\texttt{Q}_1, \texttt{Q}_2) \oplus= \texttt{reachable}_{1\bullet 2}(\texttt{P}_1, \texttt{P}_2) \otimes \texttt{edge}_1(\texttt{P}_1, \texttt{Q}_1) \otimes \texttt{edge}_2(\texttt{P}_2, \texttt{Q}_2). \quad (22)$$

Fig. 10. Four rules that, in addition to the rules in Figure 9, give the product of the two experts defined by the $\texttt{reachable}_1$ and $\texttt{reachable}_2$ predicates.

The new rules that result from this replacement can be seen in Figure 10.

### 4.2 The PRODUCT Transformation

The PRODUCT program transformation is shown in Figure 11. For each desired product of experts, where one expert, the predicate p, is defined by $n$ rules and the other expert q by $m$ rules, the transformation defines the product of experts for p•q with $n \times m$ new rules, the cross product of inference rules from the first and second experts. The value of a coupled proposition p•q in $\mathcal{P}'$ will be equal to the semiring product of p's value and q's value in $\mathcal{P}$ (or, equivalently, in $\mathcal{P}'$).

Note that lines 6–8 are nondeterministic under certain circumstances, because if the antecedent of the combined program is $\texttt{a}(\texttt{X}) \otimes \texttt{a}(\texttt{Y}) \otimes \texttt{b}(\texttt{Z})$ and the algorithm is computing the product of a and b, then the resulting antecedent could be either $\texttt{a}\bullet\texttt{b}(\texttt{X}, \texttt{Z}) \otimes \texttt{a}(\texttt{Y})$ or $\texttt{a}\bullet\texttt{b}(\texttt{Y}, \texttt{Z}) \otimes \texttt{a}(\texttt{X})$. This nondeterminism usually does not arise, and when it does, as in §5.2, there is usually an obvious preference.

The PRODUCT transformation is essentially *meaning preserving*: if the program $\mathcal{P}'$ is the result of the PRODUCT transformation on $\mathcal{P}$, then the following is true:

- Any ground instance $\texttt{p}(\mathbf{X})$ that is given a value in $\mathcal{P}$ is given the same value in $\mathcal{P}'$. This is immediately apparent because the program $\mathcal{P}'$ is stratified: none of the new rules are ever used to compute values of the form $\texttt{p}(\mathbf{X})$, so their value is identical to their value in $\mathcal{P}$.
- Any ground instance $\texttt{p}\bullet\texttt{q}(\mathbf{X}, \mathbf{Y})$ in $\mathcal{P}'$ has the same value as $\texttt{p}(\mathbf{X}) \otimes \texttt{q}(\mathbf{Y})$. This is the result of the following theorem:

**Input:** A logic program $\mathcal{P}$ and a set $\mathcal{S}$ of pairs of predicates $(\mathtt{p}, \mathtt{q})$.
**Output:** A program $\mathcal{P}'$ that extends $\mathcal{P}$, additionally computing the product predicate $\mathtt{p}\bullet\mathtt{q}$ for every pair $(\mathtt{p}, \mathtt{q}) \in \mathcal{S}$ in the input.

```
P' ← P
for all pairs (p, q) in S do
    for all rules in P, of the form p(W) ⊕= A_1 ⊗ ··· ⊗ A_n do
        for all rules in P, of the form q(X) ⊕= B_1 ⊗ ··· ⊗ B_m do
            let r ← [p•q(W, X) ⊕= A_1 ⊗ ··· ⊗ A_n ⊗ B_1 ⊗ ··· ⊗ B_m]
            for all pairs (s(Y), t(Z)) of antecedents in r such that (s, t) ∈ S do
                remove the antecedents s(Y) and t(Z) from r
                insert the antecedent s•t(Y, Z) into r
            end for
            add r to P'
        end for
    end for
end for
return P'
```

Fig. 11. Algorithmic specification of the PRODUCT transformation.

*Theorem 1*
Let $\mathcal{P}$ be a weighted logic program over a set of predicates $\mathcal{R}$, and let $\mathcal{S}$ be a set of pairs of predicates from $\mathcal{P}$. Then after applying PRODUCT on $(\mathcal{P}, \mathcal{S})$, resulting in a new program $\mathcal{P}'$, for every $(\mathtt{p}, \mathtt{q}) \in \mathcal{S}$, the value $\mathtt{p}\bullet\mathtt{q}(\mathbf{X}, \mathbf{Y})$ in $\mathcal{P}'$ is $\mathtt{p}(\mathbf{X}) \otimes \mathtt{q}(\mathbf{Y})$.

*Proof:* By distributivity of the semiring, we know that $\mathtt{p}(\mathbf{X}) \otimes \mathtt{q}(\mathbf{Y})$ is the sum: $\bigoplus_{t,r} v(t) \otimes v(r)$ where $t$ and $r$ range over proofs of $\mathtt{p}(\mathbf{X})$ and $\mathtt{q}(\mathbf{Y})$ respectively, with their values being $v(t)$ and $v(r)$. This implies that we need to show that there is a bijection between the set $A$ of proofs for $\mathtt{p}\bullet\mathtt{q}(\mathbf{X}, \mathbf{Y})$ in $\mathcal{P}'$ and the set $B$ of pairs of proofs for $\mathtt{p}(\mathbf{X})$ and $\mathtt{q}(\mathbf{Y})$ such that for every $s \in A$ and $(t, r) \in B$ we have $v(s) = v(t) \otimes v(r)$.

Using structural induction over the proofs, we first show that every pair of proofs $(t, r) \in B$ has a corresponding proof $s \in A$ with the needed value. In the base case, where the proofs $t$ and $r$ include a single step, the correspondence follows trivially. Let $(t, r) \in B$. Without loss of generality, we will assume that both $t$ and $r$ contain more than a single step in their proofs. In the last step of its proof, $t$ used a rule of the form

$$\mathtt{p}(\mathbf{X}) \oplus\!= \mathtt{a}_1(\mathbf{X}_1) \otimes \cdots \otimes \mathtt{a}_n(\mathbf{X}_n) \tag{23}$$

and $r$ used a rule in its last step of the form

$$\mathtt{q}(\mathbf{Y}) \oplus\!= \mathtt{b}_1(\mathbf{Y}_1) \otimes \cdots \otimes \mathtt{b}_m(\mathbf{Y}_m) \tag{24}$$

Let $t_i$ be the subproofs of $\mathtt{a}_i(\mathbf{X}_i)$ and let $r_j$ be subproofs of $\mathtt{b}_j(\mathbf{Y}_j)$. It follows that PRODUCT creates from those two rules a single inference rule of the form:

$$\mathtt{p}\bullet\mathtt{q}(\mathbf{X}, \mathbf{Y}) \oplus\!= \mathtt{c}_1(\mathbf{W}_1) \otimes \cdots \otimes \mathtt{c}_p(\mathbf{W}_p) \tag{25}$$

$$\texttt{reachable}_{1\bullet 2}(\texttt{Q}_1,\texttt{Q}_2) \quad \oplus\!\!= \quad \texttt{initial}_1(\texttt{Q}_1) \otimes \texttt{initial}_2(\texttt{Q}_2). \tag{26}$$

$$\texttt{reachable}_{1\bullet 2}(\texttt{Q}_1,\texttt{Q}_2) \quad \oplus\!\!= \quad \texttt{reachable}_{1\bullet 2}(\texttt{P}_1,\texttt{P}_2) \otimes \texttt{edge}_1(\texttt{P}_1,\texttt{Q}_1) \otimes \texttt{edge}_2(\texttt{P}_2,\texttt{Q}_2). \tag{27}$$

Fig. 12. By removing all but these two rules from the product of experts in Figure 10, we constrain both paths to have the same number of steps.

$$\texttt{reachable}_{1\bullet 2}(\texttt{Q}_1,\texttt{Q}_2) \oplus\!\!= \texttt{initial}_1(\texttt{Q}_1) \otimes \texttt{initial}_2(\texttt{Q}_2) \boxed{\texttt{if } \texttt{Q}_1 = \texttt{Q}_2.} \tag{28}$$

$$\texttt{reachable}_{1\bullet 2}(\texttt{Q}_1,\texttt{Q}_2) \oplus\!\!= \texttt{reachable}_{1\bullet 2}(\texttt{P}_1,\texttt{P}_2) \otimes \texttt{edge}_1(\texttt{P}_1,\texttt{Q}_1) \otimes \texttt{edge}_2(\texttt{P}_2,\texttt{Q}_2) \boxed{\texttt{if } \texttt{Q}_1 = \texttt{Q}_2.}$$

Fig. 13. By further constraining the program in Figure 12 to demand that $\texttt{Q}_1 = \texttt{Q}_2$ at all points, we constrain both paths to be identical.

$$\texttt{reachable}_{1\bullet 2}(\texttt{Q}) \quad \oplus\!\!= \quad \texttt{initial}_1(\texttt{Q}) \otimes \texttt{initial}_2(\texttt{Q}). \tag{29}$$

$$\texttt{reachable}_{1\bullet 2}(\texttt{Q}) \quad \oplus\!\!= \quad \texttt{reachable}_{1\bullet 2}(\texttt{P}) \otimes \texttt{edge}_1(\texttt{P},\texttt{Q}) \otimes \texttt{edge}_2(\texttt{P},\texttt{Q}). \tag{30}$$

Fig. 14. We can simplify Figure 13 by internalizing the side condition and giving $\texttt{reachable}_{1\bullet 2}$ only one argument.

where $\mathsf{c}_i(\mathbf{W}_i)$ is either $\mathsf{a}_k(\mathbf{Y}_k)$ for some $k$, or $\mathsf{b}_l(\mathbf{Y}_l)$ for some $k$, or $\mathsf{a}_k \bullet \mathsf{b}_\ell(\mathbf{X}_k, \mathbf{Y}_\ell)$ for some $k, \ell$.

We resolve each case as following:

1. If $\mathsf{c}_i(\mathbf{W}_i) = \mathsf{a}_k(\mathbf{Y}_k)$ then we set $s_i = t_k$.
2. If $\mathsf{c}_i(\mathbf{W}_i) = \mathsf{b}_k(\mathbf{Y}_k)$ then we set $s_i = r_k$.
3. If $\mathsf{c}_i(\mathbf{W}_i) = \mathsf{a}_k \bullet \mathsf{b}_\ell(\mathbf{X}_k, \mathbf{Y}_\ell)$ then according to the induction hypothesis, we have a proof for $\mathsf{a}_k \bullet \mathsf{b}_\ell(\mathbf{X}_k, \mathbf{Y}_\ell)$ such that its value is $v(t_k) \otimes v(r_\ell)$. We set $s_i$ to be that proof.

Since we have shown there is a proof for each antecedent of $\texttt{p} \bullet \texttt{q}(\mathbf{X}, \mathbf{Y})$, we have shown that there is a proof for $\texttt{p} \bullet \texttt{q}(\mathbf{X}, \mathbf{Y})$. That its value is indeed $\texttt{p}(\mathbf{X}) \otimes \texttt{q}(\mathbf{Y})$ is concluded trivially from the induction steps.

The reverse direction for constructing the bijection is similar, again using structural induction over proofs. □

### *4.3 From* PRODUCT *to a Product of Experts*

The output of the PRODUCT transformation is a starting point for describing dynamic programming algorithms that perform two actions—traversing a graph, scanning a string, parsing a sentence—at the same time and in a coordinated fashion. Exactly what "coordinated fashion" means depends on the problem, and answering that question determines how the problem is constrained.

If we return to the running example of generalized graph reachability, the program as written has eight rules, four from Figure 9 and four from Figure 10. Two examples of constrained product programs are given in Figures 12–14. In the first example in Figure 12, the only change is that all but two rules have been removed from the program in Figures 9 and 10. Whereas in the original product program $\texttt{reachable}_{1\bullet 2}(\texttt{Q}_1, \texttt{Q}_2)$ corresponded to the product of the weight of the best path from the initial state of graph one to $\texttt{Q}_1$ and the weight of the best path from the initial state of graph two to $\texttt{Q}_2$, the new program computes the best paths from the two origins to the two destinations with the additional requirement that the paths be the *same length*—the rules that were deleted allowed for the possibility of a prefix on one path or the other.

If our intent is for the two paths to not only have the same length but to visit exactly the same sequence of vertices, then we can further constrain the program to only define $\texttt{reachable}_{1\bullet 2}(\texttt{Q}_1, \texttt{Q}_2)$ where $\texttt{Q}_1 = \texttt{Q}_2$, as shown in Figure 13. After adding this side condition, it is no longer necessary for $\texttt{reachable}_{1\bullet 2}$ to have two arguments that are always the same, so we can simply further as shown in Figure 14. For simplicity's sake, we will usually collapse arguments that have been forced by equality constraints to agree.

The choice of paired predicates $\mathcal{S}$ is important for the final weighted logic program that PRODUCT returns and it also limits the way we can add constraints to derive a new weighted logic program. Future research might consider a machine learning setting for automatically deriving $\mathcal{S}$ from data, to minimize some cost (e.g., observed runtime). When PRODUCT is applied on two copies of the same weighted logic program (concatenated together to a single program), a natural schema for selecting paired predicates arises, in which we pair a predicate from one program with the same predicate from the other program. This "natural" pairing leads to the derivation of several useful, known algorithms, to which we turn in §5.

## 5 Examples

In this section, we give several examples of constructing weighted logic programs as constrained products of simpler weighted logic programs.

### *5.1 Finite-State Algorithms*

We have already encountered weighted finite-state automata (WSFAs) in §3.1. Like WFSAs, weighted finite-state transducers (WFSTs) are a generalization of the graph-reachability problem: in WFSAs the edges are augmented with a symbol and represented as $\texttt{arc}(\texttt{P}, \texttt{Q}, \texttt{A})$, whereas in WFSTs edges are augmented with a pair of input-output symbols and represented as $\texttt{arc}(\texttt{P}, \texttt{Q}, \texttt{A}, \texttt{B})$. Weighted finite-state machines are widely used in speech and language processing (Mohri 1997; Pereira and Riley 1997). They are used to compactly encode many competing string hypotheses, for example in speech recognition, translation, and morphological (word-structure) disambiguation. Many sequence labeling and segmentation methods can also be seen as weighted finite-state models.

$$\begin{aligned}
\texttt{goal}_{1\bullet 2} &\oplus= \texttt{path}_{1\bullet 2}(\texttt{Q}_1,\texttt{Q}_1) \otimes \texttt{final}_1(\texttt{Q}_2) \otimes \texttt{final}_2(\texttt{Q}_2). && (31)\\
\texttt{path}_{1\bullet 2}(\texttt{Q}_1,\texttt{Q}_2) &\oplus= \texttt{initial}_1(\texttt{Q}_1) \otimes \texttt{initial}_2(\texttt{Q}_2). && (32)\\
\texttt{path}_{1\bullet 2}(\texttt{Q}_1,\texttt{Q}_2) &\oplus= \texttt{path}_{1\bullet 2}(\texttt{P}_1,\texttt{P}_2) \otimes \texttt{arc}_1(\texttt{P}_1,\texttt{Q}_1,\texttt{A}_1) \otimes \texttt{arc}_2(\texttt{P}_2,\texttt{Q}_2,\texttt{A}_2)\ \boxed{\texttt{if A}_1 = \texttt{A}_2.} && (33)
\end{aligned}$$

Fig. 15. The constrained product of two of the WFSA experts described in Figure 7.

$$\begin{aligned}
\texttt{goal} &\oplus= \texttt{path}(\texttt{Q}) \otimes \texttt{final}(\texttt{Q}). && (34)\\
\texttt{path}(\texttt{Q}) &\oplus= \texttt{initial}(\texttt{Q}). && (35)\\
\texttt{path}(\texttt{Q}) &\oplus= \texttt{path}(\texttt{P}) \otimes \texttt{arc}(\texttt{P},\texttt{Q},\texttt{A},\texttt{B}). && (36)
\end{aligned}$$

Fig. 16. The weighted logic program describing weighted finite-state transducers.

$$\begin{aligned}
\texttt{goal}_{1\bullet 2} &\oplus= \texttt{path}_{1\bullet 2}(\texttt{Q}_1,\texttt{Q}_2) \otimes \texttt{final}_1(\texttt{Q}_1) \otimes \texttt{final}_2(\texttt{Q}_2). && (37)\\
\texttt{path}_{1\bullet 2}(\texttt{Q}_1,\texttt{Q}_2) &\oplus= \texttt{initial}_1(\texttt{Q}_1) \otimes \texttt{initial}_2(\texttt{Q}_2). && (38)\\
\texttt{path}_{1\bullet 2}(\texttt{Q}_1,\texttt{Q}_2) &\oplus= \texttt{path}_{1\bullet 2}(\texttt{P}_1,\texttt{P}_2) \otimes \texttt{arc}_1(\texttt{P}_1,\texttt{Q}_1,\texttt{A}_1,\texttt{B}_1) \otimes \texttt{arc}_2(\texttt{P}_2,\texttt{Q}_2,\texttt{A}_2,\texttt{B}_2)\ \boxed{\texttt{if B}_1 = \texttt{A}_2.}
\end{aligned}$$

Fig. 17. The *composition* of two weighted finite-state transducers can be derived by constraining the product of two weighted finite-state transducers.

*Weighted finite-state automata.* Our starting point for weighted finite-state automata will be the weighted logic program for WFSAs described in Figure 7, which is usually interpreted as a probabilistic automaton in the most-probable-path semiring (i.e., $\langle [0,1], \max, \times, 0, 1\rangle$). If the PRODUCT of that algorithm with itself is taken, we can follow a series of steps similar to the ones described in §4.3. First, we remove rules that would allow the two WFSAs to consider different prefixes, and then we add a constraint to rule 33 that requires the two paths' symbols to be identical. The result is a WFSA describing the *(weighted) intersection* of the two WFSAs. The intersection of two WFSAs is itself a WFSA, though it is a WFSA where states are described by two terms—$\texttt{Q}_1$ and $\texttt{Q}_2$ in $\texttt{path}_{1\bullet 2}(\texttt{Q}_1, \texttt{Q}_2)$—instead of a single term.

Weighted intersection generalizes intersection and has a number of uses. For instance, consider an FSA that is "probabilistic" but that only accepts the single string "01" because the transitions are all deterministic and have probability 1:

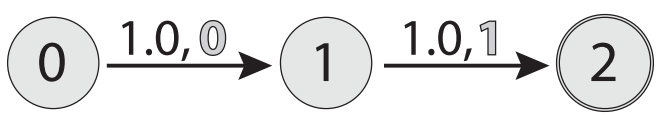

If we consider the program in Figure 15 with axioms describing the above FSA and the probabilistic FSA given in Figure 6, then the resulting program is functionally equivalent to the weighted logic program in Figure 8 describing a WFSA specialized to a particular string. Alternatively, if we consider the program in Figure 15 with axioms describing the probabilistic FSA in Figure 6 and the following single-state probabilistic FSA, the result will be a probabilistic FSA biased towards edges with the "1" symbol and against edges with the "0" symbol.

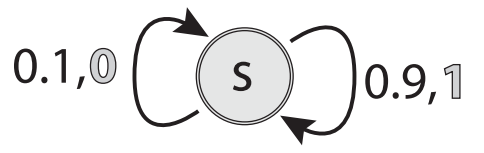

Both of the above examples can be understood as instances of the product of experts pattern discussed in §3.2. In the first case, the additional expert eliminates certain possibilities by assigning zero probability to them, and in the second case the additional expert merely modifies probabilities by preferring the symbol "1" to the symbol "0."

*Weighted finite-state transducers.* Suppose we take the PRODUCT transformation of the WFST recognition algorithm in Figure 16 with itself and constrain the result by removing all but the three interesting rules (as before) and requiring that $B_1$ (the "output" along the first edge) always be equal to $A_2$ (the "input" along the second edge). The result is shown in Figure 17; this is the recognition algorithm for the WFST resulting from *composition* of two WFSTs. Composition permits small, understandable components to be cascaded and optionally compiled, forming complex but efficient models of string transduction (Pereira and Riley 1997).

### *5.2 Context-Free Parsing*

Parsing natural languages is a difficult, central problem in computational linguistics (Manning and Schütze 1999). Consider the sentence "Alice saw Bob with binoculars." One analysis (the most likely in the real world) is that Alice had the binoculars and saw Bob through them. Another is that *Bob* had the binoculars, and Alice saw the binocular-endowed Bob. Figure 18 shows syntactic parses into noun phrases (**NP**), verb phrases (**VP**), etc., corresponding to these two meanings. It also shows some of the axioms that could be used to describe a context-free grammar describing English sentences in Chomsky normal form (Hopcroft and Ullman 1979).[5] A proof corresponds to a derivation of the given sentence in a context-free grammar, i.e., a parse tree.

Shieber et al. (1995) show that parsing with CFGs can be formalized as a logic program, and in Goodman (1999) this framework is extended to the weighted case. If weights are interpreted as probabilities, then the $\langle [0,1], \max, \times, 0, 1\rangle$ semiring interpretation finds the probability of the parse with maximum probability and the $\langle \mathbb{R}_{\geq 0} \cup \{\infty\}, +, \times, 0, 1\rangle$ semiring interpretation finds the total weight of all parse trees (a measure of the "total grammatically" of a sentence). In Figure 19, we give the specification of the weighted CKY algorithm (Cocke and Schwartz 1970; Kasami

[5] Chomsky normal form (CNF) means that the rules in the grammar are either binary with two nonterminals or unary with a terminal. We do not allow $\epsilon$ rules, which in general are allowed in CNF grammars.

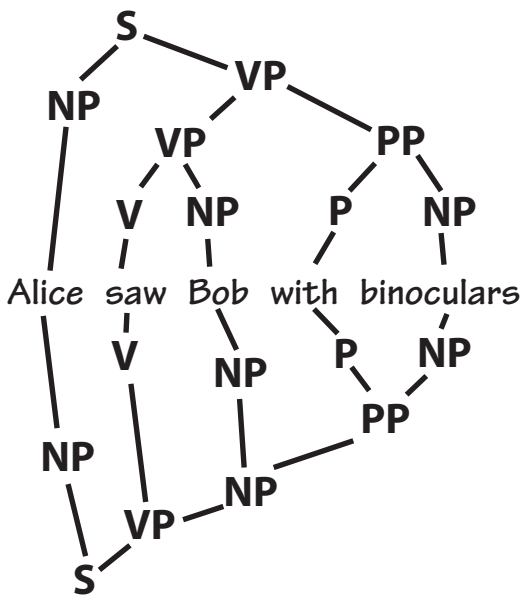

| | |
|---|---|
| **NP** → Alice | unary(np, "Alice") |
| **NP** → Bob | unary(np, "Bob") |
| **P** → with | unary(p, "with") |
| ... | ... |
| **S** → **NP VP** | binary(s, np, vp) |
| **VP** → **V NP** | binary(vp, v, np) |
| **PP** → **P NP** | binary(pp, p, np) |
| **NP** → **NP PP** | binary(np, np, pp) |
| ... | ... |

Fig. 18. An ambiguous sentence that can be parsed two ways in English (left), some of the Chomsky normal form rules for English grammar (center), and the corresponding axioms (right). There would also need to be five axioms of the form string(1, "Alice"), string(2, "saw"), etc.

$$\mathtt{goal}_1 \quad \oplus\!= \quad \mathtt{start}_1(\mathtt{S}) \otimes \mathtt{length}(\mathtt{N}) \otimes \mathtt{c}_1(\mathtt{S}, \mathtt{0}, \mathtt{N}). \tag{39}$$

$$\mathtt{c}_1(\mathtt{X}, \mathtt{I}-\mathtt{1}, \mathtt{I}) \quad \oplus\!= \quad \mathtt{unary}_1(\mathtt{X}, \mathtt{W}) \otimes \mathtt{string}(\mathtt{I}, \mathtt{W}). \tag{40}$$

$$\mathtt{c}_1(\mathtt{X}, \mathtt{I}, \mathtt{K}) \quad \oplus\!= \quad \mathtt{binary}_1(\mathtt{X}, \mathtt{Y}, \mathtt{Z}) \otimes \mathtt{c}_1(\mathtt{Y}, \mathtt{I}, \mathtt{J}) \otimes \mathtt{c}_1(\mathtt{Z}, \mathtt{J}, \mathtt{K}). \tag{41}$$

Fig. 19. A weighted logic program for parsing weighted context-free grammars.

1965; Younger 1967), which is a dynamic programming algorithm for parsing using a context-free grammar in Chomsky normal form.[6]

Figure 19 suggestively has a subscript attached to all but the length and string inputs. In our description of the product of experts framework in §3.2, the axioms length and string correspond to the conditional input sentence $I$. The unconstrained result of the PRODUCT transformation on the combination of the rules in Figure 19 and a second copy that has "2" subscripts is given in Figure 20. Under the most-probable-path probabilistic interpretation, the value of $\mathtt{goal}_{1\bullet2}$ is the probability of the given string being generated twice, once by each of the two probabilistic grammars, in each case by the most probable tree in that grammar. By constraining Figure 20, we get the more interesting program in Figure 21 that adds the additional requirement that the two parse trees in the two different grammars have the same *structure*. In particular, in all cases the constraints $\mathtt{I}_1 = \mathtt{I}_2$, $\mathtt{J}_1 = \mathtt{J}_2$, $\mathtt{K}_1 = \mathtt{K}_2$, $\mathtt{N}_1 = \mathtt{N}_2$ are added, so that instead of writing $\mathtt{c}_{1\bullet2}(\mathtt{X}_1, \mathtt{I}_1, \mathtt{J}_1, \mathtt{X}_2, \mathtt{I}_2, \mathtt{J}_2)$ we just write $\mathtt{c}_{1\bullet2}(\mathtt{X}_1, \mathtt{X}_2, \mathtt{I}, \mathtt{J})$.

*Lexicalized CFG parsing.* An interesting variant of the previous rule involves *lexicalized grammars*, which are motivated in Figure 22. Instead of describing a gram-

[6] Strictly speaking, the CKY parsing algorithm corresponds to a naïve bottom-up evaluation strategy for this program.

$$\texttt{goal}_{1\bullet 2} \;\oplus\!\!= \; \texttt{length}(\texttt{N}_1) \otimes \texttt{length}(\texttt{N}_2) \otimes \texttt{start}_1(\texttt{S}_1) \otimes \texttt{start}_2(\texttt{S}_2) \otimes \texttt{c}_{1\bullet 2}(\texttt{S}_1, 0, \texttt{N}_1, \texttt{S}_2, 0, \texttt{N}_2). \quad (42)$$

$$\texttt{c}_{1\bullet 2}(\texttt{X}_1, \texttt{I}_1 - 1, \texttt{I}_1, \texttt{X}_2, \texttt{I}_2 - 1, \texttt{I}_2) \;\oplus\!\!= \; \texttt{unary}_1(\texttt{X}_1, \texttt{W}_1) \otimes \texttt{string}(\texttt{I}_1, \texttt{W}_1) \otimes \texttt{unary}_2(\texttt{X}_2, \texttt{W}_2) \otimes \texttt{string}(\texttt{I}_2, \texttt{W}_2). \quad (43)$$

$$\texttt{c}_{1\bullet 2}(\texttt{X}_1, \texttt{I}_1 - 1, \texttt{I}_1, \texttt{X}_2, \texttt{I}_2, \texttt{K}_2) \;\oplus\!\!= \; \texttt{unary}_1(\texttt{X}_1, \texttt{W}_1) \otimes \texttt{string}(\texttt{I}_1, \texttt{W}_1) \otimes \texttt{binary}_2(\texttt{X}_2, \texttt{Y}_2, \texttt{Z}_2) \otimes \texttt{c}_2(\texttt{Y}_2, \texttt{I}_2, \texttt{J}_2) \otimes \texttt{c}_2(\texttt{Z}_2, \texttt{J}_2, \texttt{K}_2). \quad (44)$$

$$\texttt{c}_{1\bullet 2}(\texttt{X}_1, \texttt{I}_1, \texttt{K}_1, \texttt{X}_2, \texttt{I}_2 - 1, \texttt{I}_2) \;\oplus\!\!= \; \texttt{unary}_2(\texttt{X}_2, \texttt{W}_1) \otimes \texttt{string}(\texttt{I}_2, \texttt{W}_2) \otimes \texttt{binary}_1(\texttt{X}_1, \texttt{Y}_1, \texttt{Z}_1) \otimes \texttt{c}_1(\texttt{Y}_1, \texttt{I}_1, \texttt{J}_1) \otimes \texttt{c}_1(\texttt{Z}_1, \texttt{J}_1, \texttt{K}_1). \quad (45)$$

$$\texttt{c}_{1\bullet 2}(\texttt{X}_1, \texttt{I}_1, \texttt{K}_1, \texttt{X}_2, \texttt{I}_2, \texttt{K}_2) \;\oplus\!\!= \; \texttt{binary}_1(\texttt{X}_1, \texttt{Y}_1, \texttt{Z}_1) \otimes \texttt{binary}_2(\texttt{X}_2, \texttt{Y}_2, \texttt{Z}_2) \otimes \texttt{c}_{1\bullet 2}(\texttt{Y}_1, \texttt{I}_1, \texttt{J}_1, \texttt{Y}_2, \texttt{I}_2, \texttt{J}_2) \otimes \texttt{c}_{1\bullet 2}(\texttt{Z}_1, \texttt{J}_1, \texttt{K}_1, \texttt{Z}_2, \texttt{K}_2, \texttt{J}_2). \quad (46)$$

Fig. 20. The result of the PRODUCT transformation on two copies of Figure 19.

$$\texttt{goal}_{1\bullet 2} \;\oplus\!\!= \; \texttt{length}(\texttt{N}) \otimes \texttt{start}_1(\texttt{S}_1) \otimes \texttt{start}_2(\texttt{S}_2) \otimes \texttt{c}_{1\bullet 2}(\texttt{S}_1, \texttt{S}_2, 0, \texttt{N}) \quad (47)$$

$$\texttt{c}_{1\bullet 2}(\texttt{X}_1, \texttt{X}_2, \texttt{I} - 1, \texttt{I}) \;\oplus\!\!= \; \texttt{unary}_1(\texttt{X}_1, \texttt{W}) \otimes \texttt{unary}_2(\texttt{X}_2, \texttt{W}) \otimes \texttt{string}(\texttt{I}, \texttt{W}). \quad (48)$$

$$\texttt{c}_{1\bullet 2}(\texttt{X}_1, \texttt{X}_2, \texttt{I}, \texttt{K}) \;\oplus\!\!= \; \texttt{binary}_1(\texttt{X}_1, \texttt{Y}_1, \texttt{Z}_1) \otimes \texttt{binary}_2(\texttt{X}_2, \texttt{Y}_2, \texttt{Z}_2) \otimes \texttt{c}_{1\bullet 2}(\texttt{Y}_1, \texttt{Y}_2, \texttt{I}, \texttt{J}) \otimes \texttt{c}_{1\bullet 2}(\texttt{Z}_1, \texttt{Z}_2, \texttt{J}, \texttt{K}). \quad (49)$$

Fig. 21. The program in Figure 20 constrained to require internally-identical trees.

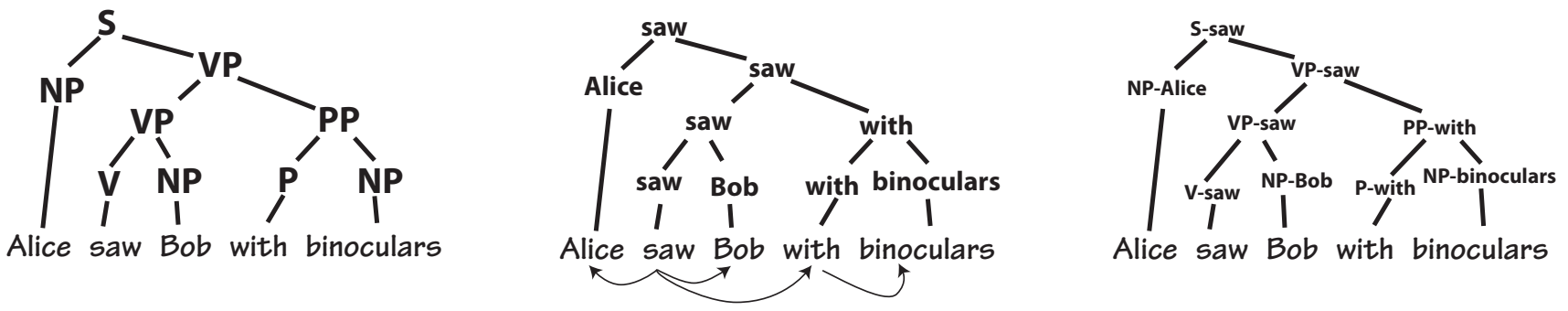

| | | |
|---|---|---|
| **NP** → Alice | **Alice** → Alice | **NP-Alice** → Alice |
| **P** → with | **with** → with | **P-with** → with |
| **S** → **NP VP** | **saw** → **Alice saw** | **S-saw** → **NP-Alice VP-saw** |
| **VP** → **V NP** | **saw** → **saw Bob** | **VP-saw** → **V-saw NP-Bob** |

Fig. 22. On the left, the grammar previously shown. In the middle, a context-free *dependency grammar*, whose derivations can be seen as parse trees (above) or a set of dependencies (below). On the right, a lexicalized grammar. Sample rules are given for each grammar.

mar using nonterminals denoting phrases (e.g., **NP** and **VP**), which is called a *constituent-structure grammar* we can define a (context-free) *dependency grammar* (Gaifman 1965) that encodes the syntax of a sentence in terms of parent-child relationships between words. In the case of the example of Figure 22, the arrows below the sentence in the middle establish "saw" as the root of the sentence; the word "saw" has three children (arguments and modifiers), one of which is the word "with," which in turn has the child "binoculars."

A simple kind of dependency grammar is a Chomsky normal form CFG where the nonterminal set is equivalent to the set of terminal symbols (so that the terminal "with" corresponds to a unique nonterminal **with**, and so on) and where all rules have the form $\mathbf{P} \rightarrow \mathbf{P}\ \mathbf{C}$, $\mathbf{P} \rightarrow \mathbf{C}\ \mathbf{P}$, and $\mathbf{W} \rightarrow w$ (where $\mathbf{P}$ is the "parent" word, $\mathbf{C}$ is the "child" word that is dependent on the parent, and $\mathbf{W}$ is the nonterminal version of terminal word $w$).

If we encode the constituent-structure grammar in the $\mathtt{unary}_1$ and $\mathtt{binary}_1$ relations and encode a dependency grammar in the $\mathtt{unary}_2$ and $\mathtt{binary}_2$ relations, then the product is a *lexicalized grammar*, like the third example from Figure 22. In particular, it describes a lexicalized context-free grammar with a product of experts probability model (Klein and Manning 2003), because the weight given to any production $\mathbf{A}\text{-}\mathbf{X} \rightarrow \mathbf{B}\text{-}\mathbf{X}\ \mathbf{C}\text{-}\mathbf{Y}$ is the semiring-product of the weight given to the production $\mathbf{A} \rightarrow \mathbf{B}\ \mathbf{C}$ and the weight given to the dependency based production $\mathbf{X} \rightarrow \mathbf{X}\ \mathbf{Y}$. This was an important distinction for Klein and Manning—they were interested in *factored* lexicalized grammars that Figure 21 can describe. These are only a small (but interesting) subset of all possible lexicalized grammars. Standard lexicalized CFGs assign weights directly to grammar productions of the form $\mathbf{A}\text{-}\mathbf{X} \rightarrow \mathbf{B}\text{-}\mathbf{X}\ \mathbf{C}\text{-}\mathbf{Y}$, not indirectly (as we do) by assigning weights to a constituent-structure and a dependency grammar. We will return to this point in §6.2 when we consider the "axiom generalization" pattern that allows us to describe general lexicalized CKY parsing (Eisner 1997; Eisner and Satta 1999).

*Nondeterminism and rule binarization.* The result of the PRODUCT transformation shown in Figure 20 was the first time the nondeterminism inherent in lines 6-8 of the description of the PRODUCT transformation (Figure 11) has come into play. Because there were two $\mathtt{c}_1$ premises and two $\mathtt{c}_2$ premises, they could have been merged in more than one way. For example, the following would have been a potential alternative to rule 46:

$$\begin{aligned} \mathtt{c}_{1\bullet 2}(\mathtt{X}_1, \mathtt{I}_1, \mathtt{K}_1, \mathtt{X}_2, \mathtt{I}_2, \mathtt{K}_2) \quad \oplus= \quad & \mathtt{binary}_1(\mathtt{X}_1, \mathtt{Y}_1, \mathtt{Z}_1) \otimes \mathtt{binary}_2(\mathtt{X}_2, \mathtt{Y}_2, \mathtt{Z}_2) \ \otimes \\ & \mathtt{c}_{1\bullet 2}(\mathtt{Y}_1, \mathtt{I}_1, \mathtt{J}_1, \mathtt{Z}_2, \mathtt{K}_2, \mathtt{J}_2) \otimes \mathtt{c}_{1\bullet 2}(\mathtt{Z}_1, \mathtt{J}_1, \mathtt{K}_1, \mathtt{Y}_2, \mathtt{I}_2, \mathtt{J}_2). \end{aligned} \tag{50}$$

However, this would have broken the correspondence between $\mathtt{I}_1$ and $\mathtt{I}_2$ and made it impossible to constrain the resulting program as we did. An alternative to CKY is the *binarized* variant of CKY where rule 41 is split into two rules by introducing a new, temporary predicate (rules 51 and 52):

$$\mathtt{temp}_1(\mathtt{X}, \mathtt{Y}, \mathtt{J}, \mathtt{K}) \quad \oplus= \quad \mathtt{binary}_1(\mathtt{X}, \mathtt{Y}, \mathtt{Z}) \otimes \mathtt{c}_1(\mathtt{Z}, \mathtt{J}, \mathtt{K}). \tag{51}$$

$$\mathtt{c}_1(\mathtt{X}, \mathtt{I}, \mathtt{K}) \quad \oplus= \quad \mathtt{c}_1(\mathtt{Y}, \mathtt{I}, \mathtt{J}) \otimes \mathtt{temp}_1(\mathtt{X}, \mathtt{Y}, \mathtt{J}, \mathtt{K}). \tag{52}$$

In this variant, the nondeterministic choice in the PRODUCT transformation disappears. The choice that we made in pairing was consistent with the choice that is forced in the binarized CKY program.

$$\texttt{goal} \;\oplus\!\!= \; \texttt{targetlength}(\texttt{M}) \otimes \texttt{predict}(\texttt{E}_{\texttt{M}-1}, \texttt{E}_{\texttt{M}}, \texttt{M}+1). \quad (53)$$

$$\texttt{predict}(\texttt{E}_{\texttt{J}-1}, \texttt{E}_{\texttt{J}}, \texttt{J}+1) \;\oplus\!\!= \; \texttt{predict}(\texttt{E}_{\texttt{J}-2}, \texttt{E}_{\texttt{J}-1}, \texttt{J}) \otimes \texttt{trigram}(\texttt{E}_{\texttt{J}-2}, \texttt{E}_{\texttt{J}-1}, \texttt{E}_{\texttt{J}}). \quad (54)$$

Fig. 23. A weighted logic program giving a trigram prediction model for a language, which can be generalized to an $n$-gram model for any $n$.

$$\texttt{goal} \;\oplus\!\!= \; \texttt{sourcelength}(\texttt{N}) \otimes \texttt{trans}(\texttt{N}, []). \quad (55)$$

$$\texttt{trans}(\texttt{I}', \texttt{Es}) \;\oplus\!\!= \; \texttt{trans}(\texttt{I}, []) \otimes \texttt{phrase}(\texttt{I}, \texttt{I}', \texttt{E}_{\texttt{J}} :: \texttt{Es}). \quad (56)$$

$$\texttt{trans}(\texttt{I}', \texttt{Es}) \;\oplus\!\!= \; \texttt{trans}(\texttt{I}', \texttt{E}_{\texttt{J}} :: \texttt{Es}). \quad (57)$$

Fig. 24. A weighted logic program that describes monotone decoding—translating a phrase at a time of the input language into the output language without reordering.

$$\texttt{goal} \;\oplus\!\!= \; \texttt{sourcelength}(\texttt{N}) \otimes \texttt{targetlength}(\texttt{M}) \otimes \texttt{pr}\bullet\texttt{tr}(\texttt{N}, \texttt{M}, \texttt{E}_{\texttt{M}-1}, \texttt{E}_{\texttt{M}}, \texttt{M}+1). \quad (58)$$

$$\texttt{pr}\bullet\texttt{tr}(\texttt{I}', \texttt{J}+1, \texttt{E}_{\texttt{J}-1}, \texttt{E}_{\texttt{J}}, \texttt{Es}) \;\oplus\!\!= \; \texttt{pr}\bullet\texttt{tr}(\texttt{I}, \texttt{J}, \texttt{E}_{\texttt{J}-2}, \texttt{E}_{\texttt{J}-1}, []) \otimes \texttt{trigram}(\texttt{E}_{\texttt{J}-2}, \texttt{E}_{\texttt{J}-1}, \texttt{E}_{\texttt{J}}) \otimes \texttt{phrase}(\texttt{I}, \texttt{I}', \texttt{E}_{\texttt{J}} :: \texttt{Es}). \quad (59)$$

$$\texttt{pr}\bullet\texttt{tr}(\texttt{I}', \texttt{J}+1, \texttt{E}_{\texttt{J}-1}, \texttt{E}_{\texttt{J}}, \texttt{Es}) \;\oplus\!\!= \; \texttt{pr}\bullet\texttt{tr}(\texttt{I}', \texttt{J}, \texttt{E}_{\texttt{J}-2}, \texttt{E}_{\texttt{J}-1}, \texttt{E}_{\texttt{J}} :: \texttt{Es}) \otimes \texttt{trigram}(\texttt{E}_{\texttt{J}-2}, \texttt{E}_{\texttt{J}-1}, \texttt{E}_{\texttt{J}}). \quad (60)$$

Fig. 25. Phrase translation as the constrained product of Figures 23 and 24.

### *5.3 Translation Algorithms*

Another example of two probabilistic models that play the role of experts arises in translation of sentences from one natural language to another. We will summarize how the PRODUCT transformation was applied to a simple form of phrase-to-phrase translation (Koehn et al. 2003) by Lopez (2009).

Lopez (2009) suggested a deductive view of algorithms for machine translation, similar to the view of parsing given in Shieber et al. (1995). Lopez used the PRODUCT transformation to derive an algorithm for phrase translation from two different factor programs, one which attempts to enforce *fluency* (a measure of the grammaticality of a sentence) in the translated sentence and one which attempts to enforce *adequacy* (a measure of how much of the meaning of an original sentence is preserved in the translation.)

If fluency is a measure of the grammaticality of a sentence, then it would seem that the CKY algorithm for parsing context-free grammars would be a candidate. While such models have been used in translation (Charniak et al. 2003), Lopez's example uses a simpler notion of fluency based on an $n$-gram language model (Manning and Schütze 1999, Chapter 6). An $n$-gram model assigns the probability of a sentence to be the product of probabilities of each word following the $(n-1)$-word sequence immediately preceding it. As a concrete example, let us say that $n = 3$ (called a "trigram" model) and work with the program in Figure 23. If we were estimating our trigram model based on the relative frequencies of sequences in Shakespeare's *Othello*, we would note that the phrase "if it" appears eight times in the text. Three of these are from the sequence "if it be" and one is from the sequence "if it prove," so the axiom $\texttt{trigram}(\text{``}\texttt{if}\text{''}, \text{``}\texttt{it}\text{''}, \text{``}\texttt{be}\text{''})$ should have a probability that is three times the probability given to $\texttt{trigram}(\text{``}\texttt{if}\text{''}, \text{``}\texttt{it}\text{''}, \text{``}\texttt{prove}\text{''})$. If we then stared the program with the initial sentence fragment $\texttt{predict}(\text{``}\texttt{if}\text{''}, \text{``}\texttt{it}\text{''}, 3)$, we could derive $\texttt{predict}(\text{``}\texttt{it}\text{''}, \text{``}\texttt{be}\text{''}, 4)$ with the aforementioned axiom and then $\texttt{predict}(\text{``}\texttt{be}\text{''}, \text{``}\texttt{demanded}\text{''}, 5)$ with the axiom $\texttt{trigram}(\text{``}\texttt{it}\text{''}, \text{``}\texttt{be}\text{''}, \text{``}\texttt{demanded}\text{''})$, a sequence occurring once in the text. The result so far is a sequence "if it be demanded" that does not appear in *Othello*, but which perhaps sounds like it could (which is an informal way of describing the criterion for fluency).

The weighted logic program Lopez uses to enforce adequacy is the "monotone decoding" logic program presented in Figure 24. The program is slightly contrived in order to interact with the PRODUCT transformation correctly. The atomic proposition $\texttt{trans}(\texttt{I}, \texttt{Es})$ refers to a particular point, $\texttt{I}$, in the source-language string and a list $\texttt{Es}$ of unprocessed words in the target language.[7] Each deduction consumes a single word ($\texttt{E}_\texttt{J}$) in the target language—indeed, this is the only function of rule 57. When there are no words to remove, then either the entire source-language string has been translated (rule 55), or else progress can continue by translating some chunk of the source-language sentence starting from position $\texttt{I}$ and ending at position $\texttt{I}'$ as the non-empty list of target-language words $\texttt{E}_\texttt{J} :: \texttt{Es}$ and applying rule 56. This translation of a sequence of the source-language words is captured by the premise $\texttt{phrase}(\texttt{I}, \texttt{I}', \texttt{Es})$, corresponding to the source subsequence from position $\texttt{I}$ to position $\texttt{I}'$ being translated as $\texttt{Es}$ (a target-language phrase). The meaning of $\texttt{phrase}$ could be defined by a set of axioms or by a rule. In the latter case, if we enumerate all the substrings $\texttt{Ds}$ in the source-language sentence as axioms $\texttt{substr}(\texttt{I}, \texttt{I}', \texttt{Ds})$ and provide axioms $\texttt{ptranslate}(\texttt{Ds}, \texttt{Es})$ describing source-language to target-language phrase translation, then $\texttt{phrase}(\texttt{I}, \texttt{I}', \texttt{Es})$ may be defined by the following rule:

$$\texttt{phrase}(\texttt{I}, \texttt{I}', \texttt{Es}) \quad \oplus\!= \quad \texttt{substr}(\texttt{I}, \texttt{I}', \texttt{Ds}) \otimes \texttt{ptranslate}(\texttt{Ds}, \texttt{Es}). \tag{61}$$

Note that $\texttt{substr}$ might be provided as an axiom, or derived from axioms encoding the source sentence through another inference rule.

Figure 25 displays Lopez's phrase translation program by constraining the product of the $n$-gram model and monotone decoding programs. Lopez describes this for any $n$, but for simplicity we continue using a trigram model ($n = 3$). The combined predicate simultaneously tracks a position in the source-language sentence $\texttt{I}$ and the target-language sentence $\texttt{J}$. The word $\texttt{E}_\texttt{J}$ that was discarded at each step in Figure 24 is given relevance by the trigram model. The combination of these two programs uses the monotone decoding program's capabilities to make sure that the phrase-by-phrase meaning of the source-language string $D_1 \ldots D_N$ is preserved in the destination language string $E_1 \ldots E_M$ (adequacy) while simultaneously using the trigram model's capabilities to ensure that the result is a plausible sentence in the destination language (fluency).

[7] We use a standard syntactic shorthand for lists; "[]" can be read as the constant $\texttt{nil}$ and "$\texttt{E} :: \texttt{Es}$" can be read as $\texttt{cons}(\texttt{E}, \texttt{Es})$.

Our presentation of machine translation algorithms through the PRODUCT transformation is simplistic. Lopez (2009) discusses more powerful translation algorithms that permit, for example, reordering of phrases.

## 6 Variations on PRODUCT

Up to this point, we have viewed our use of the PRODUCT transformation as one that solves a problem of *joint optimization*: we take two logic programs that describe structures (such as strings, paths, or trees), relate them to one another by adding constraints, and then optimize over the two original structures simultaneously (one instance of this is when we use weighted logic programming to describe a product of experts.) This is a useful pattern, but it is not the only interesting use of the fold/unfold transformation underlying the PRODUCT transformation. In this section we consider two other variants: in the first we only optimize over one of the two structures and fix the other one, and in the second we take the output of PRODUCT as describing not joint optimization over two simple structures but over one complex structure.

### *6.1 Fixing One of the Factor Structures*

The usual use of the PRODUCT transformation is to joint optimization on two structures, but general side conditions can be used to take the additional step of *fixing* one of the two structures and having the weighted logic program perform optimization on the *other* structure, subject to constraints imposed through the pairing.

In the setting where we consider weights to be probabilities, this is useful for solving certain probabilistic inference problems. Using the path-sum semiring (i.e., $\langle \mathbb{R}_{\geq 0} \cup \{\infty\}, +, \times, 0, 1\rangle$), the result is a program calculating the *marginalized quantity* $p(x) = \sum_y p(x, y)$ (where $x$ corresponds to one program's proof and $y$ to the other program's proof). This is a useful quantity in learning; for example, the expectation-maximization (EM) algorithm (Dempster et al. 1977) for optimizing the marginalized log-likelihood of observed structures requires calculating *sufficient statistics* which are based on marginal quantities. Using the most-probable-path semiring (i.e., $\langle [0, 1], \max, \times, 0, 1\rangle$), the result is a program for solving $\text{argmax}_y\, p(y \mid x)$—that is, for finding the most probable $y$ given the fixed $x$.

The transformation of the constrained result of the PRODUCT transformation to a program with one proof fixed is essentially mechanical. We consider the example of lexicalized parsing from Figure 22. We take the constituent-structure parse as the structure we want to fix in order optimize over the possible matching parses from the dependency grammar. The shape of the constituent-structure parse tree can be represented by a series of new axioms that mirror the structure of the $\mathtt{c_1(X, I, J)}$ pred-

$$\texttt{path}_{1\bullet 2}(\texttt{Q}_1,\texttt{Q}_2) \;\oplus\!= \; \texttt{initial}_{1\bullet 2}(\texttt{Q}_1,\texttt{Q}_2). \tag{63}$$

$$\texttt{path}_{1\bullet 2}(\texttt{Q}_1,\texttt{Q}_2) \;\oplus\!= \; \texttt{path}_{1\bullet 2}(\texttt{P}_1,\texttt{P}_2) \otimes \texttt{arc}_{1\bullet 2}(\texttt{P}_1,\texttt{P}_2,\texttt{Q}_1,\texttt{Q}_2,\texttt{A}_1,\texttt{A}_2). \tag{64}$$

Fig. 26. Weighted finite-state transducers as the product of two weighted finite-state machines.

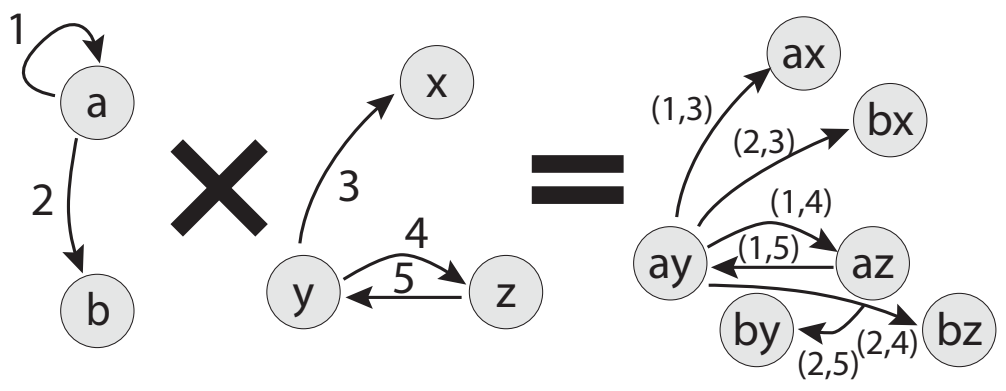

Fig. 27. A finite-state transducer that can be expressed as the PRODUCT of two finite-state automata.

icate defining the constituent-structure grammar: $\texttt{proof}_1(\texttt{s},0,5)$, $\texttt{proof}_1(\texttt{np},0,1)$, $\texttt{proof}_1(\texttt{vp},1,5)$, $\texttt{proof}_1(\texttt{vp},1,3)$, $\texttt{proof}_1(\texttt{pp},3,5)$, and so on.

Then we take the constrained PRODUCT of CKY that we used to describe lexicalized parsing (Figure 21) and, wherever there was a conclusion derived from $\texttt{c}_1$, we add a matching side condition that references $\texttt{proof}_1$. The critical rule (49) ends up looking like this:

$$\begin{aligned}\texttt{c}_{1\bullet 2}(\texttt{X}_1,\texttt{X}_2,\texttt{I},\texttt{K}) \;\oplus\!=\; & \texttt{binary}_1(\texttt{X}_1,\texttt{Y}_1,\texttt{Z}_1) \otimes \texttt{binary}_2(\texttt{X}_2,\texttt{Y}_2,\texttt{Z}_2) \otimes \\ & \texttt{c}_{1\bullet 2}(\texttt{Y}_1,\texttt{Y}_2,\texttt{I},\texttt{J}) \otimes \texttt{c}_{1\bullet 2}(\texttt{Z}_1,\texttt{Z}_2,\texttt{J},\texttt{K}) \texttt{ if } \texttt{proof}_1(\texttt{X}_1,\texttt{I},\texttt{K}).\end{aligned} \tag{62}$$

The effect of this additional constraint is to disqualify any proof that does not match the constituent-structure grammar which we have fixed and encoded as $\texttt{proof}_1$ axioms. The idea of partially constraining CFG derivations with some bracketing structure was explored by Pereira and Schabes (1992).

### *6.2 Axiom Generalization*

Axiom generalization is another way of manipulating products of weighted logic programs in a way that reveals the simple structures underlying a complex structure. Figure 26, which is intended to describe a weighted finite-state transducer, is close to the weighted logic program in Figure 15 that describes the intersection of two finite-state machines, but there are two differences. First, we have *not* forced the two symbols to be the same; instead, we wish to interpret $\texttt{A}_1$ from the first expert as the transducer's input symbol and $\texttt{A}_2$ as the transducer's output symbol. Second, we have merged $\texttt{initial}_1(\texttt{Q}_1) \otimes \texttt{initial}_2(\texttt{Q}_2)$ to the single product predicate $\texttt{initial}_{1\bullet 2}(\texttt{Q}_1,\texttt{Q}_2)$, and likewise for $\texttt{arc}$. As a first approximation, we can just define $\texttt{arc}_{1\bullet 2}$ (and, similarly, $\texttt{initial}_{1\bullet 2}$) by a single rule of this form:

$$\texttt{arc}_{1\bullet 2}(\texttt{P}_1,\texttt{P}_2,\texttt{Q}_1,\texttt{Q}_2,\texttt{A}_1,\texttt{A}_2) \oplus\!= \texttt{arc}_1(\texttt{P}_1,\texttt{Q}_1,\texttt{A}_1) \otimes \texttt{arc}_2(\texttt{P}_2,\texttt{Q}_2,\texttt{A}_2) \qquad (65)$$

An example is given in Figure 27. Two finite-state *machines*, one with two states (a and b) and one with three states (x, y, and z), are shown—we are working over the Boolean semiring, so each arc in the figure corresponds to a true-valued arc axiom. The PRODUCT of these two experts in the manner of Figure 26 is a single finite-state *transducer* with six states.

However, we can only describe a certain subset of finite-state transducers as the direct product of finite-state machines in this way. If we consider all possible Boolean-valued finite-state transducers with two symbols and one state, we have 16 possible transducers, but only 10 that can be "factored" as two independent finite-state machines, such as these three:

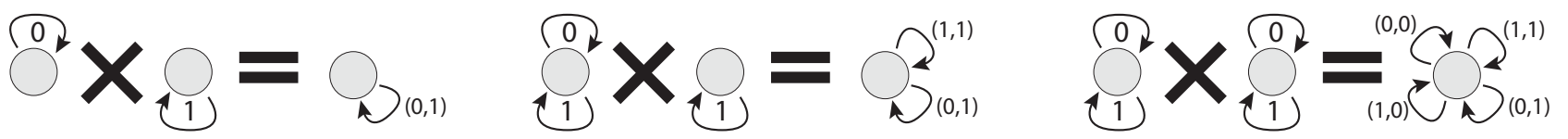

Six others, like the NOT transducer that outputs 1 given the input 0 and outputs 0 given the input 1, cannot be represented as the product of two FSMs.

In many settings, limiting ourselves to the "factorable" finite-state transducers (or lexicalized grammars) can have conceptual or computational advantages. When this does not suffice, we can perform *axiom generalization*, which amounts to removing the requirement of Eq. 65 that the value of atomic propositions of the form $\texttt{arc}_{1\bullet 2}$ be the product of an atomic proposition of the form $\texttt{arc}_1$ and an atomic proposition of the form $\texttt{arc}_2$. If we directly define axioms of the form $\texttt{arc}_{1\bullet 2}$, we can describe transducers in their full generality.

This represents a new way of thinking about the PRODUCT transformation. Thus far, we have considered the result of the PRODUCT transformation as a way of describing programs that work over two different structures. Axiom generalization suggests that we can consider the PRODUCT transformation as a way of taking two programs that work over individual structures and deriving a new program that works over a *single more complicated structure* that, in special cases, can be factored into two different structures. This is particularly relevant in the area of lexicalized grammars and parsing where the general, more complicated structure is what came first and the factored models which we have considered thus far arose later as special cases.

*Parsing algorithms and the* PRODUCT *transformation.* Many parsing algorithms can be derived by using the PRODUCT transformation as a way of deriving programs that do not neatly factor into two parts. Lexicalized parsing is a simple example; Figure 28 derives a lexicalized parser by performing axiom generalization on Figure 21. The grammar production "**P-with** $\rightarrow$ with" can be represented by including the axiom $\texttt{unary}_{1\bullet 2}(\texttt{p}, \text{“}\texttt{with}\text{”})$, and the binary production **S-saw** $\rightarrow$ **NP-Alice VP-saw** can be represented by the axiom $\texttt{binary}_{1\bullet 2}(\texttt{s}, \text{“}\texttt{saw}\text{”}, \texttt{np}, \text{“}\texttt{alice}\text{”}, \texttt{vp}, \text{“}\texttt{saw}\text{”})$.

Synchronous grammars are another instance in which the axiom generalization

$$\mathtt{goal}_{1\bullet 2} \oplus= \mathtt{length}(\mathtt{N}) \otimes \mathtt{start}(\mathtt{S}) \otimes \mathtt{c}_{1\bullet 2}(\mathtt{S},\mathtt{W},0,\mathtt{N}). \quad (66)$$
$$\mathtt{c}_{1\bullet 2}(\mathtt{X},\mathtt{W},\mathtt{I}-1,\mathtt{I}) \oplus= \mathtt{unary}_{1\bullet 2}(\mathtt{X},\mathtt{W}) \otimes \mathtt{string}(\mathtt{I},\mathtt{W}). \quad (67)$$
$$\mathtt{c}_{1\bullet 2}(\mathtt{X},\mathtt{W},\mathtt{I},\mathtt{K}) \oplus= \mathtt{binary}_{1\bullet 2}(\mathtt{X},\mathtt{W},\mathtt{Y},\mathtt{W}_1,\mathtt{Z},\mathtt{W}_2) \otimes \mathtt{c}_{1\bullet 2}(\mathtt{Y},\mathtt{W}_1,\mathtt{I},\mathtt{J}) \otimes \mathtt{c}_{1\bullet 2}(\mathtt{Z},\mathtt{W}_2,\mathtt{J},\mathtt{K}). \quad (68)$$

Fig. 28. A algorithm for CKY over a general lexicalized grammar derived from Figure 21 by axiom generalization.

$$\mathtt{goal} \oplus= \mathtt{length}(\mathtt{N}) \otimes \mathtt{start}(\mathtt{S}) \otimes \mathtt{c}(\mathtt{S},0,\mathtt{N}). \quad (69)$$
$$\mathtt{c}(\mathtt{X},\mathtt{I},\mathtt{I}) \oplus= \mathtt{unary}(\mathtt{X},\epsilon) \otimes \mathtt{pos}(\mathtt{I}). \quad (70)$$
$$\mathtt{c}(\mathtt{X},\mathtt{I}-1,\mathtt{I}) \oplus= \mathtt{unary}(\mathtt{X},\mathtt{W}) \otimes \mathtt{string}(\mathtt{I},\mathtt{W}). \quad (71)$$
$$\mathtt{c}(\mathtt{X},\mathtt{I},\mathtt{K}) \oplus= \mathtt{binary}(\mathtt{X},\mathtt{Y},\mathtt{Z}) \otimes \mathtt{c}(\mathtt{Y},\mathtt{I},\mathtt{J}) \otimes \mathtt{c}(\mathtt{Z},\mathtt{J},\mathtt{K}). \quad (72)$$

Fig. 29. A variant of CKY that handles grammar productions of the form $\mathbf{X} \rightarrow \epsilon$.

$$\mathtt{goal}_{1\bullet 2} \oplus= \mathtt{length}_1(\mathtt{M}) \otimes \mathtt{length}_2(\mathtt{N}) \otimes \mathtt{start}_{1\bullet 2}(\mathtt{S}) \otimes \mathtt{c}_{1\bullet 2}(\mathtt{S},0,\mathtt{N},0,\mathtt{M}). \quad (73)$$
$$\mathtt{c}_{1\bullet 2}(\mathtt{X},\mathtt{I}-1,\mathtt{I},\mathtt{J},\mathtt{J}) \oplus= \mathtt{unary}_{1\bullet 2}(\mathtt{X},\mathtt{W}_1,\epsilon) \otimes \mathtt{string}_1(\mathtt{I},\mathtt{W}_1) \otimes \mathtt{pos}_2(\mathtt{J}). \quad (74)$$
$$\mathtt{c}_{1\bullet 2}(\mathtt{X},\mathtt{I},\mathtt{I},\mathtt{J}-1,\mathtt{J}) \oplus= \mathtt{unary}_{1\bullet 2}(\mathtt{X},\epsilon,\mathtt{W}_2) \otimes \mathtt{pos}_1(\mathtt{I}) \otimes \mathtt{string}_2(\mathtt{J},\mathtt{W}_2). \quad (75)$$
$$\mathtt{c}_{1\bullet 2}(\mathtt{X},\mathtt{I}-1,\mathtt{I},\mathtt{J}-1,\mathtt{J}) \oplus= \mathtt{unary}_{1\bullet 2}(\mathtt{X},\mathtt{W}_1,\mathtt{W}_2) \otimes \mathtt{string}_1(\mathtt{I},\mathtt{W}_1) \otimes \mathtt{string}_2(\mathtt{J},\mathtt{W}_2). \quad (76)$$
$$\mathtt{c}_{1\bullet 2}(\mathtt{X}_1,\mathtt{I}_1,\mathtt{K}_1,\mathtt{I}_2,\mathtt{K}_2) \oplus= \mathtt{binary}_{1\bullet 2}(\mathtt{X},\mathtt{Y},\mathtt{Z}) \otimes \mathtt{c}_{1\bullet 2}(\mathtt{Y},\mathtt{I}_1,\mathtt{J}_1,\mathtt{I}_2,\mathtt{J}_2) \otimes \mathtt{c}_{1\bullet 2}(\mathtt{Z},\mathtt{J}_1,\mathtt{K}_1,\mathtt{J}_2,\mathtt{K}_2). \quad (77)$$

Fig. 30. A simple transduction grammar derived from Figure 29.

$$\mathtt{c}_{1\bullet 2}(\mathtt{X}_1,\mathtt{I}_1,\mathtt{K}_1,\mathtt{I}_2,\mathtt{K}_2) \oplus= \mathtt{inversion}_{1\bullet 2}(\mathtt{X},\mathtt{Y},\mathtt{Z}) \otimes \mathtt{c}_{1\bullet 2}(\mathtt{Y},\mathtt{I}_1,\mathtt{J}_1,\mathtt{J}_2,\mathtt{K}_2) \otimes \mathtt{c}_{1\bullet 2}(\mathtt{Z},\mathtt{J}_1,\mathtt{K}_1,\mathtt{I}_2,\mathtt{J}_2). \quad (78)$$

Fig. 31. By adding to Figure 30 these rule corresponding to the *other* way that the $\mathtt{c}_1$ and $\mathtt{c}_2$ antecedents may be merged in the PRODUCT transformation, we can describe an inversion transduction grammar.

view is interesting. A synchronous grammar can be thought of as parsing two different sentences in two different languages with two different grammars using a *single* parse tree. For example, if $\mathbf{X} \rightarrow \mathbf{YZ}$ is a grammar production in one language and $\mathbf{A} \rightarrow \mathbf{BC}$ is a grammar production in another language, then $\mathbf{X\text{-}A} \rightarrow \mathbf{Y\text{-}B}\ \mathbf{Z\text{-}C}$ is a possible grammar production in the synchronous grammar.

A *transduction grammar* (Wu 1997), is a synchronous grammar which generates two isomorphic derivations with a trivial alignment between the nodes of those two derivations. We can describe a parser for a transduction grammar with the program in Figure 30. Synchronous grammars need to be able to deal with situations in which a word in one language does not appear in the matching sentence in the other language; this is done by starting from the enriched CKY program in Figure 29 that can handle grammar productions of the form $\mathbf{X} \to \epsilon$.

In practice, transduction grammars do a bad job of aligning two sentences in different natural languages that are translations of each other, because it is often the case that two parts of a pair of sentences need to be in opposite positions relative to one another—in language one, the verb phrase might precede a prepositional phrase, and in language two, the corresponding verb phrase might *follow* the corresponding prepositional phrase. An *inversion transduction grammar* describes an alternate form of grammar production, which Wu (1997) writes as $\mathbf{X} \to \langle \mathbf{YZ} \rangle$. This grammar production declares that if $A_1$ and $A_2$ simultaneously parse as $\mathbf{Y}$ in languages one and two (respectively) and $B_1$ and $B_2$ simultaneously parse as $\mathbf{Z}$ in languages one and two (respectively), then $A_1 B_1$ and $B_2 A_2$ simultaneously parse as $\mathbf{Z}$.

Somewhat surprisingly, this inversion production rule can be described using the *alternate* allowable way of merging the premises when the PRODUCT transformation is performed on two copies of the CKY algorithm, as discussed in §5.2 (see rule 50). By adding this alternate form as given in Figure 31, we can describe the algorithm for parsing with inversion transduction grammars described by Wu (1997).

## 7 The Entropy Semiring and Kullback-Leibler Divergence

An important construct in information theory and machine learning is the Kullback-Leibler (KL) divergence (Kullback and Leibler 1951). KL divergence is a function of two probability distributions over the same event space. It measures their dissimilarity, though it is not, strictly speaking, a distance (it is not symmetric). For two distributions $p$ and $q$ for random variable $X$ ranging over events $x \in \mathcal{X}$, KL divergence is defined as

$$\mathrm{KL}(p\|q) = \sum_{x\in\mathcal{X}} p(X=x)\log\frac{p(X=x)}{q(X=x)} \quad (79)$$

$$= \underbrace{\sum_{x\in\mathcal{X}} p(X=x)\log p(X=x)}_{-H(p)} - \underbrace{\sum_{x\in\mathcal{X}} p(X=x)\log q(X=x)}_{\mathrm{CE}(p\|q)} \quad (80)$$

where $H(p)$ denotes the Shannon entropy of the distribution $p$ (Shannon 1948), a measure of uncertainty, and $\mathrm{CE}(p\|q)$ denotes the cross-entropy between $p$ and $q$.[8] A full discussion of these information-theoretic quantities is out of scope for this paper; we note that they are widely used in statistical machine learning (Koller and Friedman 2009). In this section, we first show how the entropy of $p(P)$, with $P$ ranging over proofs of `goal` (the axioms corresponding to random variables $A$ and $I$ are suppressed here, for clarity), can be calculated using a weighted logic program, following Hwa (2004). We then describe a generalization of a result of Cortes et al. (2006) to show how to use PRODUCT to produce a weighted logic program for calculating the KL divergence between the two distributions induced by the WLPs.

[8] In brief, the Shannon entropy of distribution $p$ is the expected number of bits required to send a message drawn according to $p$ under an optimal coding scheme. Cross-entropy is the average number of bits required to encode a message in the optimal coding scheme for $q$ when messages are actually distributed according to $p$. Hence $\mathrm{KL}(p\|q) = \mathrm{CE}(p\|q) - H(p)$ is the average number of *extra* bits required when the true distribution of messages is $p$ but the coding scheme is based on $q$. Note that $\mathrm{KL}(p\|p) = 0$. If there is an event $x \in \mathcal{X}$ such that $p(x) > 0$ and $q(x) = 0$, then $\mathrm{KL}(p\|q) = +\infty$.

### *7.1 Generalized Entropy Semiring*

The domain of the generalized entropy semiring is $(\mathbb{R} \cup \{+\infty, -\infty\})^3$. The multiplication and addition operations are defined as follows:

$$\langle x_1, y_1, z_1 \rangle \oplus \langle x_2, y_2, z_2 \rangle \quad = \quad \langle x_1 + x_2, y_1 + y_2, z_1 + z_2 \rangle \tag{81}$$

$$\langle x_1, y_1, z_1 \rangle \otimes \langle x_2, y_2, z_2 \rangle \quad = \quad \langle x_1 x_2, x_1 y_2 + x_2 y_1, z_1 z_2 \rangle \tag{82}$$

These operations have the required closure, associativity, and commutativity properties previously discussed for semirings. See Cortes et al. (2006) for a proof which can be extended trivially to our generalized semiring.

Suppose we have a weighted logic program such that the path-sum (in the $\langle \mathbb{R}_{\geq 0} \cup \{\infty\}, +, \times, 0, 1 \rangle$ semiring) is 1 (i.e., the value of the `goal` theorem is 1). If we map the weights of all axioms in the original program to new values in the generalized entropy semiring, we can use the new semiring to calculate the Shannon entropy of the distribution over proofs of `goal`:

$$-\sum_{proof} p(P = proof) \log p(P = proof) \tag{83}$$

where $x$ ranges over proofs of `goal`. The mapping is simply $w \mapsto \langle w, -w \log w, 0 \rangle$. (The third element of the semiring value is not needed here.) If we solve the new weighted logic program and achieve value $\langle w', h', 0 \rangle$ for the `goal` theorem, then under our assumption that $w' = 1$ (the value of `goal` in the original program in the real semiring), $h'$ is the entropy of the distribution over the proof random variable (given the axioms and `goal`). The formal result is given as a corollary in §7.2.

This semiring can be used, for example, with the CKY algorithm from Figure 19. It makes the derivation of the tree entropy for context-free grammars (i.e., the entropy over the context-free derivations for an ambiguous string) automatic, and obviates the design of a specific algorithm for computing the tree entropy for probabilistic context-free grammars, as described in Hwa (2004). With the CKY algorithm, a proof *proof* in Eq. 83 represents a derivation in the grammar. Similarly, a weighted logic program describing a finite-state transducer (Figure 16) can be used to compute the entropy of hidden sequences for hidden Markov models as described by Hernando et al. (2005).

We now relax the assumption that the sum of all proof scores is 1. Suppose that the value of the `goal` theorem in the generalized entropy semiring is $(w', h', 0)$, with $w' \neq 1$. In this case, $h'$ is not the entropy of a proper probability distribution.

We can renormalize the scores of the proofs, $u(\mathit{proof})$, by dividing by $w'$, treating them as a proper conditional distribution (conditioning on the truth of the `goal` theorem); then the entropy of this conditional distribution, $\frac{u(\mathit{proof})}{w'}$, is

$$
\begin{aligned}
& -\sum_{\mathit{proof}} \frac{u(\mathit{proof})}{w'} \log \frac{u(\mathit{proof})}{w'} && (84)\\
&= \frac{1}{w'}\left(-\sum_{\mathit{proof}} u(\mathit{proof})(\log u(\mathit{proof}) - \log w')\right)\\
&= \frac{1}{w'}\left(h' + (\log w') \sum_{\mathit{proof}} u(\mathit{proof})\right) = \frac{1}{w'}\,(h' + w' \log w') = \frac{h'}{w'} + \log w'
\end{aligned}
$$

Therefore, whenever we can use weighted logic programming (in the real semiring) to calculate sums of proof scores, we can use the generalized entropy semiring to find the Shannon entropy of the (possibly renormalized) distribution over proofs. The renormalization uses $w'$ and $h'$, two quantities that are calculated directly when we use the generalized entropy semiring.

### *7.2 KL Divergence Between Proof Distributions and* PRODUCT

Cortes et al. (2006) showed how to compute the KL divergence (also called relative entropy) between two distributions over strings defined by probabilistic FSA, using a construct similar to our generalized entropy semiring. We generalize that result to KL divergence over two *proof* distributions $p(P)$ and $q(P)$ given by a weighted logic program $\mathcal{P}$. We assume in this discussion that the set of axioms with non-zero weights are identical under $p$ and under $q$; the general setting where this does not hold is correctly handled, using $a \log 0 = -\infty$ and $0 \log a = 0$ for all $a > 0$.

We abuse notation slightly and use $p$ and $q$ to denote the values of axioms, theorems, and proofs in the real semiring weighted logic programs used to calculate the sum of proof-scores for `goal` under axioms weighted according to $p$ and $q$. Let $\mathrm{Proofs}(t)$ denote the set of logical proofs of a theorem $t$, and for $x \in \mathrm{Proofs}(t)$, let $p(t)$—respectively, $q(t)$—denote the score of the proof $x$:

$$p(t) = \sum_{x \in \mathrm{Proofs}(t)} p(x) \quad (85)$$

$$q(t) = \sum_{x \in \mathrm{Proofs}(t)} q(x) \quad (86)$$

$$(87)$$

We seek the KL divergence:

$$\mathrm{KL}(p \| q) = \sum_{x \in \mathrm{Proofs}(\mathtt{goal})} p(x) \log \frac{p(x)}{q(x)} \quad (88)$$

In order to accomplish this calculation, we will first map the weights of axioms under $p$ and $q$ into the generalized entropy semiring as follows, for any axiom $a$:

$$\langle p(a), q(a)\rangle \mapsto \langle p(a), p(a)\log q(a), q(a)\rangle \quad (89)$$

For a theorem $t$, let

$$R(t) = \sum_{x\in\text{Proofs}(t)} p(x)\log q(x) \quad (90)$$

*Theorem 2*
Solving $\mathcal{P}$ in the generalized entropy semiring with weights defined as above results in goal having value $\langle p(\texttt{goal}), R(\texttt{goal}), q(\texttt{goal})\rangle$.

*Proof:* We will treat the weighted logic program as a set of equations with all left-hand-side variables grounded. We will use upper-case to refer to free variables (e.g., $\mathbf{Z} = \langle Z_1, \ldots\rangle$) and lower-case to refer to grounded values (e.g., $\mathbf{z} = \langle z_1, \ldots\rangle$). The range of values that variables $\mathbf{Z}$ can get is denoted by $\text{Rng}(\mathbf{Z})$. The weighted logic program can be seen as a set of equations:

$$\texttt{c}(\mathbf{w}) = \bigoplus_{[\texttt{c}(\mathbf{w}) \mathrel{\oplus}= \texttt{a}_\texttt{i}(\mathbf{w}',\mathbf{Z})\otimes\texttt{b}_\texttt{i}(\mathbf{w}'',\mathbf{Z})]\in\mathcal{P},\mathbf{w}'\subseteq\mathbf{w},\mathbf{w}''\subseteq\mathbf{w}} \bigoplus_{\mathbf{z}\in\text{Rng}(\mathbf{Z})} \texttt{a}_\texttt{i}(\mathbf{w}',\mathbf{z}) \otimes \texttt{b}_\texttt{i}(\mathbf{w}'',\mathbf{z}) \quad (91)$$

(Note that any of $\mathbf{w}$, $\mathbf{w}'$, $\mathbf{w}''$, and $\mathbf{z}$ may be empty.)

We now show that the value achieved for $\texttt{c}(\mathbf{w})$ when solving in the semiring is

$$\langle p(\texttt{c}(\mathbf{w})), \sum_{x\in\text{Proofs}(\texttt{c}(\mathbf{w}))} p(x)\log q(x), q(\texttt{c}(\mathbf{w}))\rangle \quad (92)$$

where $\text{Proofs}(\texttt{c}(\mathbf{w}))$ denotes the set of proofs for $\texttt{c}(\mathbf{w})$. We will show that the solution of Equations 91 is the value in Equation 92 for $\texttt{c}(\mathbf{w})$.

For the first and third coordinates, this equality follows naturally because of the definition of the generalized entropy semiring: the first and third coordinates are equivalent to the non-negative real semiring used for summing over proof scores under the two value assignments $p$ and $q$, respectively.

Consider a particular $\oplus$-addend to the value of $\texttt{c}(\mathbf{w})$,

$$\begin{aligned}
&\texttt{a}_\texttt{i}(\mathbf{w}',\mathbf{z}) \otimes \texttt{b}_\texttt{i}(\mathbf{w}'',\mathbf{z}) && (93)\\
&= \langle p(\texttt{a}_\texttt{i}(\mathbf{w}',\mathbf{z})), R(\texttt{a}_\texttt{i}(\mathbf{w}',\mathbf{z})), q(\texttt{a}_\texttt{i}(\mathbf{w}',\mathbf{z}))\rangle \\
&\quad \otimes \langle p(\texttt{b}_\texttt{i}(\mathbf{w}'',\mathbf{z})), R(\texttt{b}_\texttt{i}(\mathbf{w}'',\mathbf{z})), q(\texttt{b}_\texttt{i}(\mathbf{w}'',\mathbf{z}))\rangle && (94)\\
&= \left\langle \begin{array}{l} p(\texttt{a}_\texttt{i}(\mathbf{w}',\mathbf{z}))p(\texttt{b}_\texttt{i}(\mathbf{w}'',\mathbf{z})), \\ p(\texttt{a}_\texttt{i}(\mathbf{w}',\mathbf{z}))R(\texttt{b}_\texttt{i}(\mathbf{w}'',\mathbf{z})) + p(\texttt{b}_\texttt{i}(\mathbf{w}'',\mathbf{z}))R(\texttt{a}_\texttt{i}(\mathbf{w}',\mathbf{z})), \\ q(\texttt{a}_\texttt{i}(\mathbf{w}',\mathbf{z})))q(\texttt{b}_\texttt{i}(\mathbf{w}'',\mathbf{z})) \end{array} \right\rangle && (95)
\end{aligned}$$

Consider the second coordinate.

$$\begin{aligned}
&p(\texttt{a}_\texttt{i}(\mathbf{w}',\mathbf{z}))R(\texttt{b}_\texttt{i}(\mathbf{w}'',\mathbf{z})) + p(\texttt{b}_\texttt{i}(\mathbf{w}'',\mathbf{z}))R(\texttt{a}_\texttt{i}(\mathbf{w}',\mathbf{z})) && (96)\\
&= \left( p(\texttt{a}_\texttt{i}(\mathbf{w}',\mathbf{z})) \sum_{x\in\text{Proofs}(\texttt{b}_\texttt{i}(\mathbf{w}'',\mathbf{z}))} p(x)\log q(x) \right)
\end{aligned}$$

$$+ \left( p(\mathtt{b_i}(\mathbf{w}'', \mathbf{z})) \sum_{x' \in \text{Proofs}(\mathtt{a_i}(\mathbf{w}', \mathbf{z}))} p(x') \log q(x') \right) \tag{97}$$

$$= \left( \sum_{x' \in \text{Proofs}(\mathtt{a_i}(\mathbf{w}', \mathbf{z}))} p(x') \sum_{x \in \text{Proofs}(\mathtt{b_i}(\mathbf{w}'', \mathbf{z}))} p(x) \log q(x) \right)$$

$$+ \left( \sum_{x \in \text{Proofs}(\mathtt{b_i}(\mathbf{w}'', \mathbf{z}))} p(x) \sum_{x' \in \text{Proofs}(\mathtt{a_i}(\mathbf{w}', \mathbf{z}))} p(x') \log q(x') \right) \tag{98}$$

$$= \sum_{x \in \text{Proofs}(\mathtt{b_i}(\mathbf{w}'', \mathbf{z}))} \sum_{x' \in \text{Proofs}(\mathtt{a_i}(\mathbf{w}', \mathbf{z}))} p(x) p(x') \log(q(x) q(x')) \tag{99}$$

Embedding the above in a $\oplus$-summation over $\mathbf{z}$ and a $\oplus$-summation over inference rule instantiations gives a $\oplus$-summation over proofs of $\mathtt{c}(\mathbf{w})$,

$$\sum_{x \in \text{Proofs}(\mathtt{c}(\mathbf{w}))} p(x) \log q(x) \tag{100}$$

which is $R(\mathtt{c}(\mathbf{w}))$ as desired. □

Denote by $(\bar{p}, \bar{R}, \bar{q})$ the value for $\mathtt{goal}$ in the generalized entropy semiring as discussed above, i.e., $\bar{p} = p(\mathtt{goal})$, $\bar{R} = R(\mathtt{goal})$, and $\bar{q} = q(\mathtt{goal})$. If we wish to renormalize $p$ by $\bar{p}$ and $q$ by $\bar{q}$ to give proper distributions over proofs of $\mathtt{goal}$ (given axioms and $\mathtt{goal}$), then

$$\text{CE}\left( \frac{1}{\bar{p}} p \middle\| \frac{1}{\bar{q}} q \right) = \frac{\bar{R}}{\bar{p}} - \log \bar{q} \tag{101}$$

Noting that $-H(p) = \text{CE}(p \| p)$,

$$\text{KL}\left( \frac{1}{\bar{p}} p \middle\| \frac{1}{\bar{q}} q \right) = \text{CE}(p \| p) - \text{CE}(p \| q) \tag{102}$$

we can solve for the KL divergence of two (possibly renormalized) distributions $p$ and $q$ using the above results. Alternatively, if the generalized KL divergence between *unnormalized* distributions is preferred (O'Sullivan 1998), note that (in the notation of the above):

$$\sum_{x \in \text{Proofs}(\mathtt{goal})} \left( p(x) \log \frac{p(x)}{q(x)} - p(x) + q(x) \right) = \bar{R} - \bar{p} + \bar{q} \tag{103}$$

Cortes et al. describe how to compute KL divergence between two probabilistic finite-state automata with a single path per string ("unambiguous" automata). The authors make use of finite-state intersection (discussed above in §5.1). This suggests an analogous interpretation of the PRODUCT transformation for computing KL divergence between two weighted logic programs.

Let $\mathcal{P}$ and $\mathcal{Q}$ be two instances of a weighted logic program, with possibly different different axiom weights. Assume we set the values of the axioms of $\mathcal{P}$ (ranging over $a$) to be $\langle p(a), 0, 1 \rangle$, and for $\mathcal{Q}$ we set them to $\langle 1, \log q(a), q(a) \rangle$. If we take a PRODUCT of $\mathcal{P}$ and $\mathcal{Q}$, using the "natural" pairing, then we end up with a program that computes $\langle p(\mathtt{goal}), R(\mathtt{goal}), q(\mathtt{goal}) \rangle$ in the generalized entropy semiring, where

$$\texttt{goal} \oplus= \texttt{path}(\texttt{P},\texttt{Q}) \otimes \texttt{final}(\texttt{Q}). \quad (104)$$
$$\texttt{path}(\texttt{null},\texttt{Q}) \oplus= \texttt{initial}(\texttt{Q}). \quad (105)$$
$$\texttt{path}(\texttt{P}',\texttt{Q}) \oplus= \texttt{path}(\texttt{P},\texttt{P}') \otimes \texttt{biarc}(\texttt{P},\texttt{P}',\texttt{Q},\texttt{A},\texttt{B}). \quad (106)$$

Fig. 32. Weighted finite-state transducer where arriving at a certain state depends on the last two states. `null` serves as a place-holder for the non-state prior to the initial state.

$$\texttt{goal}_{1\bullet 2} \oplus= \texttt{path}_{1\bullet 2}(\texttt{P},\texttt{Q}) \otimes \texttt{final}_1(\texttt{Q}) \otimes \texttt{final}_2(\texttt{Q}). \quad (107)$$
$$\texttt{path}_{1\bullet 2}(\text{null},\texttt{Q}) \oplus= \texttt{initial}_1(\texttt{Q}) \otimes \texttt{initial}_2(\texttt{Q}). \quad (108)$$
$$\texttt{path}_{1\bullet 2}(\texttt{P}',\texttt{Q}) \oplus= \texttt{path}_{1\bullet 2}(\texttt{P},\texttt{P}') \otimes \texttt{arc}_1(\texttt{P}',\texttt{Q},\texttt{A},\texttt{B}) \otimes \texttt{biarc}_2(\texttt{P},\texttt{P}',\texttt{Q},\texttt{A},\texttt{B}) \quad (109)$$

Fig. 33. The PRODUCT program of Figure 16 with Figure 32, with constraints that match proofs according to states and emissions sequences.

$R(\cdot)$ is specified in Eq. 90. These quantities can be used to compute KL divergence as specified in Eq. 102. This is a direct result of Theorem 2.

### *7.3 KL Divergence and Projections*

We can use PRODUCT to calculate KL divergence between proof distributions even when $\mathcal{P}$ and $\mathcal{Q}$ are not two instances of the same program. We consider cases where the proofs of $\mathcal{P}$ and the proofs of $\mathcal{Q}$ have a shared semantics, that is, each proof of either $\mathcal{P}$ or $\mathcal{Q}$ maps to an event in some "interpretation space."

As an example, consider the WLP in Figure 16 describing a weighted finite-state transducer. In a more general formulation, where each state depends on the previous $N$ states visited, rather than just the single most recent state. This modification is reflected in Figure 32 for $N = 2$. The axiom $\texttt{biarc}(\texttt{P}',\texttt{Q},\texttt{P},\texttt{A},\texttt{B})$ is to be interpreted as: "if the last two states were `Q` and $\texttt{P}'$, transfer to state `P` while reading symbol `A` and emitting the symbol `B`." Since the two programs have different axioms, the spaces of their respective proofs are different. However, both programs have identical semantics to a proof: a proof (in either program) corresponds to a sequence of states that the transducers go through together with the reading of a symbol and the emission of another symbol.

Running PRODUCT on the WFST in Figure 16 (we call it $\mathcal{P}$) and the WFST in Figure 32 (we call it $\mathcal{Q}$) with a particular pairing and constraints (such that the paths are identical) yields the program in Figure 33. If we let the axioms $a$ in $\mathcal{P}$ have the values $\langle p(a), 0, 1\rangle$ and the axioms $a$ in $\mathcal{Q}$ the values $\langle 1, \log q(a), q(a)\rangle$, then the resulting PRODUCT program in Figure 33, as implied by Theorem 2, calculates the KL divergence between two distributions over the set of state paths: one which is defined using a finite-state transducer with $N = 1$ and the other with $N = 2$.

We now generalize this idea for two different programs $\mathcal{P}$ and $\mathcal{Q}$. We assume that PRODUCT is applied in such a way that axioms from $\mathcal{P}$ are paired only with axioms from $\mathcal{Q}$, and vice versa. Further, each proof in the PRODUCT program must decompose into exactly one proof in $\mathcal{P}$ and one proof in $\mathcal{Q}$.[9] For a proof in the PRODUCT program, $y$, we define $\pi_{\mathcal{P}}(y)$ ($\pi_{\mathcal{Q}}(y)$) to be the projection of $y$ to a proof in $\mathcal{P}$ ($\mathcal{Q}$). The "projection" of a proof is a separation of the proof which uses coupled theorems and axioms into theorems and axioms of only one of the programs. For example, projecting a proof $y$ in the product program in Figure 33 yields two proofs: $\pi_{\mathcal{P}}(y)$ describes a sequence of transitions through the transducer with $N = 1$ and $\pi_{\mathcal{Q}}(y)$ describes a sequence of transitions through the transducer with $N = 2$; yet both proofs correspond to the same sequence of states.

In the generalized entropy semiring, we set the values of the axioms of $\mathcal{P}$ to be $\langle p(a), 0, 1\rangle$, and for $\mathcal{Q}$ we set them to $\langle 1, \log q(a), q(a)\rangle$. The PRODUCT program computes $\langle p(\texttt{goal}_1), R(\texttt{goal}_{1\cdot 2}), q(\texttt{goal}_2)\rangle$. This time, the summation in $R(\texttt{goal}_{1\cdot 2})$ is over proofs which are implicitly paired:

$$R(\texttt{goal}_{1\cdot 2}) \quad = \sum_{y \in \text{Proofs}(\texttt{goal}_{1\cdot 2})} p(\pi_{\mathcal{P}}(y)) \log q(\pi_{\mathcal{Q}}(y)) \tag{110}$$

The quantities $p(\texttt{goal}_1)$, $R(\texttt{goal}_{1\cdot 2})$, and $q(\texttt{goal}_2)$ can be used as before to compute the KL divergence between the distributions over the shared "interpretation space" of the proofs in the two programs. This technique is only correct when interpretations are in a one-to-one correspondence with the proofs in $\mathcal{P}$ and with the proofs in $\mathcal{Q}$, and PRODUCT is applied so that equivalently-interpretable proofs in the two programs are paired.

We note that in the general case, the problem of computing KL divergence between two arbitrary distributions is hard. For example, with Markov networks, there are restrictions, which resemble the restrictions we pose, of clique decomposition (Koller and Friedman 2009).

## 8 Conclusion

We have described a framework for dynamic programming algorithms whose solutions correspond to proof values in two constrained weighted logic programs. Our framework includes a program transformation, PRODUCT, which combines the two weighted logic programs that compute over two structures into a single weighted logic program for a joint proof. Appropriate constraints, encoded intuitively as variable unification or side conditions in the weighted logic program, are then added manually. The framework naturally captures and permits generalization of many existing algorithms. We have shown how variations on the the program transformation enable to include a larger set of algorithms as the result of the program transformation. We have concluded by showing how the program transformation can be used to interpret the computation of Kullback-Leibler divergence for two weighted logic programs which are defined over an identical interpretation space.

[9] Note that these constraints are satisfied in the case of two identical programs with the "natural" pairing, as in §7.2.

## Acknowledgments

The authors acknowledge helpful comments from the two anonymous reviewers, Jason Eisner, Rebecca Hwa, Adam Lopez, Alberto Pettorossi, Frank Pfenning, Sylvia Rebholz and David Smith. This research was supported by an NSF graduate fellowship to the second author and NSF grant IIS-0713265 and an IBM faculty award to the third author.